\documentclass[11pt]{article}

\usepackage[preprint]{acl}

\usepackage{times}
\usepackage{latexsym}

\usepackage[T1]{fontenc}
\usepackage{amsmath}
\usepackage{url}
\usepackage[utf8]{inputenc}

\usepackage{microtype}
\usepackage{booktabs}
\usepackage{tabularx}
\usepackage{inconsolata}

\usepackage{graphicx}
\usepackage{CJKutf8}
\usepackage{enumitem}
\usepackage{multirow}
\newcommand\newfootnote[1]{%
  \begingroup
  \renewcommand\thefootnote{}\footnote{#1}%
  \addtocounter{footnote}{-1}%
  \endgroup
}

\usepackage[most]{tcolorbox}
\definecolor{promptbox_color}{HTML}{9DC8C8}
\definecolor{promptbox_color_2}{HTML}{a5d296}
\usepackage[table]{xcolor}  
\usepackage{colortbl}   
\usepackage{fontawesome5}
\title{ChinaHeritaQA: A Culturally-Grounded Visual Question Answering Dataset for World Heritage Sites in China}

\author{
  \textbf{Yi Zhang\textsuperscript{*1,2}}~~~
  \textbf{Bolei Ma\textsuperscript{*1,3}}~~~
    \textbf{Yong Cao\textsuperscript{4}}~~~
    \textbf{Chengyan Wu\textsuperscript{5}}~~~
\\\vspace{5pt}
\textbf{Daniel Hershcovich\textsuperscript{6}}~~~
\textbf{Anna-Carolina Haensch\textsuperscript{1,3,7}}
\\
  \textsuperscript{1}LMU Munich  \textsuperscript{2}FAU Erlangen-Nuremberg \textsuperscript{3}Munich Center for Machine Learning \\
  \textsuperscript{4}University of Tübingen \& Tübingen AI Center 
    \textsuperscript{5}Sun Yat-sen University 
  \\%\vspace{5pt}
    \textsuperscript{6}University of Copenhagen   \textsuperscript{7}University of Maryland, College Park 
}

\begin{document}
\maketitle
\begin{abstract}

We introduce \textbf{ChinaHeritaQA}, a multimodal benchmark dataset for evaluating the cultural reasoning abilities of vision-language models (VLMs) on UNESCO World Heritage sites in China. The dataset comprises 2,279 in-the-wild images paired with 14,133 bilingual (Chinese/English) multiple-choice QA pairs spanning seven cognitive dimensions, from basic identity recognition to historical periodization and architectural analysis. Guided by a UNESCO-aligned heritage ontology and verified through rigorous human annotation, the dataset ensures linguistic quality and factual consistency. 
Evaluations of state-of-the-art VLMs reveal that while top models exceed human performance on average, substantial task-level variation emerges: models excel at visual recognition but struggle with culturally grounded reasoning. Performance also varies by dynasty and region. ChinaHeritaQA reveals that strong visual retrieval does not extend to cultural and historical understanding. We release the dataset to support future research on culturally aware multimodal learning.
\newfootnote{$^\ast$Equal contributions. Contact: 
bolei.ma@lmu.de}

\textbf{Resources:} \\
{\small 
\faDatabase \ \href{https://huggingface.co/datasets/Multilingual-NLP/ChinaHeritaQA}{Multilingual-NLP/ChinaHeritaQA} \\
\faGithub \ \href{https://github.com/boleima/ChinaHeritaQA}{boleima/ChinaHeritaQA}}

\end{abstract}

\section{Introduction}

Recent vision-language models (VLMs) have shown impressive capabilities across a wide range of multimodal tasks~\cite{liu2023mmbench,fu2025mme,li2023seed}, yet the benchmarks used to evaluate them are predominantly built from Western or English-centric data~\cite{liu-etal-2021-visually,yin2021broaden}. This creates a systematic gap when models encounter non-Western visual and cultural content \cite{NEURIPS2024_9a16935b}, where understanding an image often requires integrating historical knowledge, regional symbolism, and cultural context rather than just identifying common objects.

Cultural heritage sites pose a particular challenge for current VLMs. Unlike everyday scenes, heritage images carry layers of meaning tied to specific historical periods, architectural traditions, and regional identities. General evaluation suites such as VQAv2~\cite{goyal2017making}, MMBench~\cite{liu2023mmbench}, SEED-Bench~\cite{li2023seed}, and MMMU~\cite{yue2024mmmu,yue-etal-2025-mmmu} cover broad perceptual and reasoning skills but do not test the cultural and historical knowledge that heritage images demand. Chinese-specific benchmarks such as CMMMU~\cite{zhang2024cmmmuchinesemassivemultidiscipline} and CVLUE~\cite{Wang_Liu_Yu_Huang_Li_Wan_Che_Chen_2025} include cultural content but treat it as a subset of general encyclopedic knowledge rather than a structured reasoning domain.

\begin{figure}[t]
    \centering
    \includegraphics[width=1\linewidth]{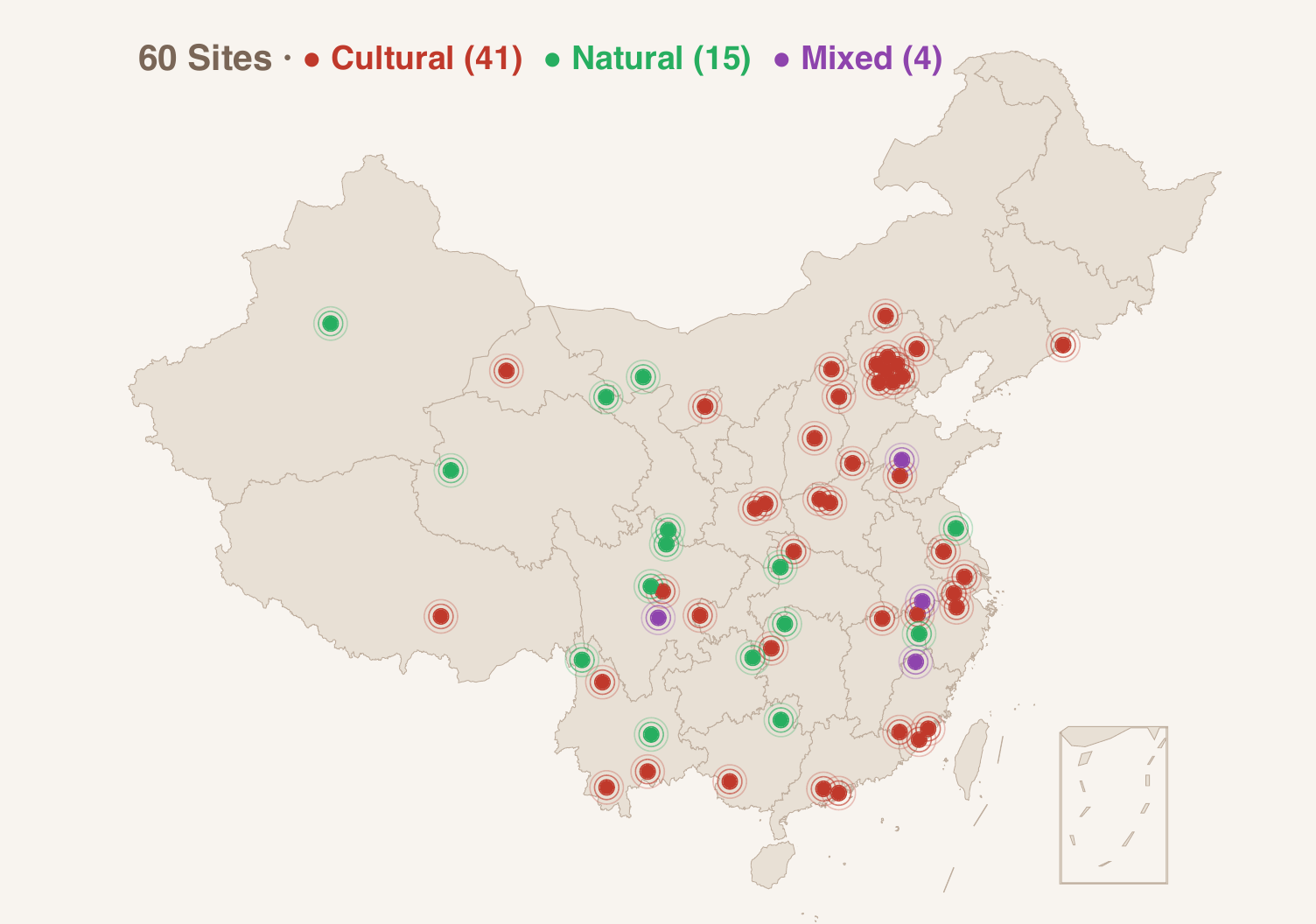}
    \caption{The distribution of World Cultural Heritage Sites in China according to UNESCO, including Cultural, Natural and Mixed Heritage Sites.}
    \label{fig:china}
\end{figure}

China provides a natural focus for this line of research. As of 2026, it holds 60 UNESCO World Heritage sites, one of the highest counts globally, spanning more than 5,000 years of architectural and cultural history from Neolithic earthworks and Tang-dynasty grottoes to Ming imperial palaces and Qing garden complexes.\footnote{\url{https://whc.unesco.org/}.} Figure \ref{fig:china} shows the distribution of these sites in China. Heritage tourism is a major sector worldwide \cite{richards2018cultural, timothy2003heritage}, and Chinese heritage sites attract hundreds of millions of visitors annually. These sites vary widely in style, period, material, and function, and because they are primarily documented and interpreted in Chinese, they also present a bilingual dimension that existing benchmarks have not addressed.

We introduce \textbf{ChinaHeritaQA}, a bilingual multimodal benchmark for evaluating VLMs on Chinese World Heritage. The dataset consists of 2,279 images collected paired with 14,133 multiple-choice QA pairs in Chinese and English, covering Chinese UNESCO-defined heritage sites. Drawing images from social media rather than encyclopedic archives reflects how visitors actually see and document these places~\cite{urry2011tourist}: under varied lighting, from different angles, and at different distances. Questions span seven dimensions, ranging from site identification and visual grounding to historical periodization, functional analysis, and architectural reasoning.

We evaluate six open-weight VLMs against a human performance baseline with native Chinese speakers. Top-performing models exceed human accuracy on most question types, with the widest advantage on site recognition and visual grounding. However, this aggregate result masks considerable task-level variation: performance drops substantially on questions requiring historical periodization, functional analysis, and architectural knowledge, revealing that current VLMs are better at visual retrieval than at grounding images in domain-specific cultural and historical knowledge. Model performance also varies by dynasty and geographic region, reflecting the uneven coverage of heritage knowledge in pretraining data.

Our contributions are as follows:
\begin{itemize}[leftmargin=*,itemsep=1pt, topsep=2pt]
    \item We introduce \textbf{ChinaHeritaQA}, the first large-scale bilingual VQA benchmark for Chinese UNESCO World Heritage, comprising 2,279 in-the-wild images and 14,133 multiple-choice QA pairs across 7 cognitive dimensions.
\item We design a structured annotation pipeline combining a UNESCO-aligned heritage ontology, LLM-assisted QA generation, cross-cultural distractors, and rigorous human verification.
    \item We evaluate state-of-the-art VLMs alongside a human performance baseline, revealing persistent gaps in historical and culturally grounded reasoning.
    \item We provide fine-grained analyses by question type, dynasty, and region, identifying specific failure modes and directions for future work.
\end{itemize}

\section{Related Work}

\paragraph{General Vision-Language Evaluation.} The rapid evolution of Large VLMs has driven the development of comprehensive evaluation suites like MMBench \cite{liu2023mmbench}, MME \cite{fu2025mme}, SEED-Bench \cite{li2023seed}, and MMMU \cite{yue2024mmmu}. While these benchmarks assess wide-ranging perceptual and reasoning tasks, they primarily rely on data from the English-speaking web or generic object datasets. Consequently, they often fail to capture visual semantics specific to non-Western regions, leading to severe performance gaps when models encounter culturally dense imagery \cite{liu-etal-2021-visually}.

\paragraph{Cultural Bias and Geo-diversity in VLMs.} A growing body of literature highlights the ``Western-centric'' bias in multimodal benchmarks \cite{NEURIPS2024_1568882b}. \citet{liu-etal-2021-visually} and GeoDE \cite{yin2021broaden} pioneered benchmarks using native concepts to demonstrate that state-of-the-art models struggle with geo-diverse visual reasoning. Recent works have expanded this scope to regional domains, such as food culture (FoodieQA \cite{li-etal-2024-foodieqa}, WorldCuisines \cite{winata-etal-2025-worldcuisines}) and Southeast Asian nuances \cite{satar-etal-2025-seeing}, and art-critique \cite{yu2026vulcabench}. In digital heritage, datasets like Artpedia \cite{stefanini2019artpedia} connect artwork with text but focus heavily on Western art. Our work extends this line of research to Chinese cultural heritage, where the interaction between history, architecture, and bilingual semantics poses unique, unaddressed challenges.

\paragraph{Visual Benchmarks for Chinese Culture.} Chinese-specific multimodal benchmarks have recently emerged, including CMMMU \cite{zhang2024cmmmuchinesemassivemultidiscipline} and CVLUE \cite{Wang_Liu_Yu_Huang_Li_Wan_Che_Chen_2025}. However, they treat cultural knowledge as a general encyclopedic subdomain rather than a focused domain requiring structured historical reasoning. Other specialized datasets target narrow subfields like traditional clothing \cite{zhou-etal-2025-hanfu},  calligraphy \cite{yang-etal-2025-recontextualizing}, or cultural artifacts \cite{10.1007/978-3-032-04627-7_33}. 
No existing benchmark targets World Heritage. ChinaHeritaQA fills this gap by combining in-the-wild visual diversity with structured historical and architectural reasoning.

\paragraph{Heritage Documentation and Tourism Analytics.} Tourism and landscape studies show that heritage perception is fundamentally visual and culturally mediated \cite{urry2011tourist,richards2018cultural}, where architectural styles and spatial arrangements define site authenticity \cite{daniel2001whither}. Social media has further transformed how visitors document and share these historic environments \cite{giaccardi2012heritage}. However, diverse cultural material remains challenging for both vision and language models \cite{hershcovich-etal-2022-challenges}. Our ChinaHeritaQA dataset is a valuable contribution to evaluation in this important field.

\begin{figure*}[t]
    \centering
    \includegraphics[width=\textwidth]{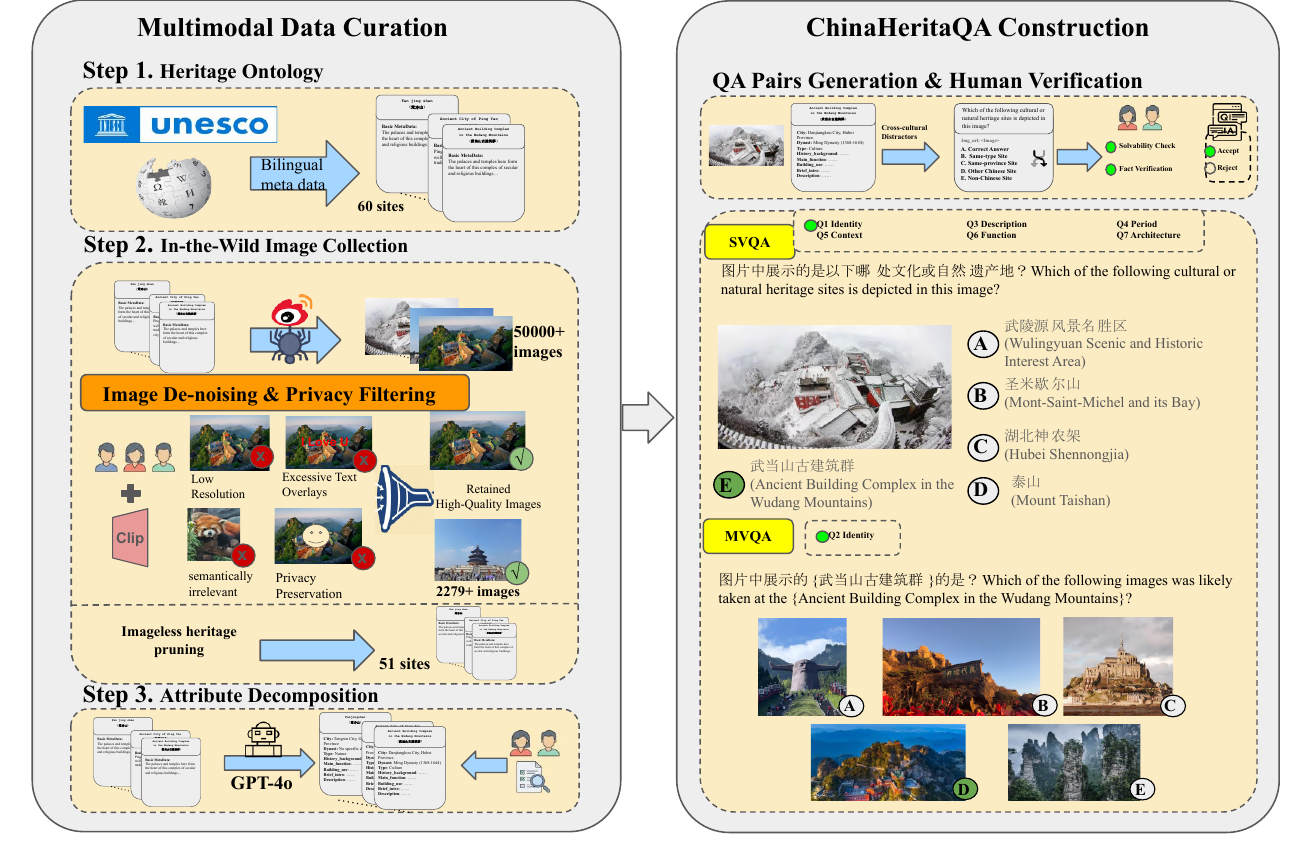}
    \caption{The overall construction pipeline of ChinaHeritaQA. The framework consists of two main phases. \textbf{Left:} Multimodal Data Curation Pipeline. %: We construct a heritage ontology based on UNESCO standards (Step 1) to guide the collection of large-scale in-the-wild images from social media (Step 2). After rigorous image de-noising and privacy filtering (Step 3), GPT-4o is employed to transform unstructured bilingual textual descriptions into structured site-level attributes, (Step 4). 
    \textbf{Right:} ChinaHeritaQA Construction \& Validation. %: Based on the decomposed attributes, we generate QA pairs equipped with cross-cultural distractors to prevent shortcut learning. These candidates undergo strict human verification for solvability and factual accuracy. 
    The final benchmark evaluates VLMs through SVQA and MVQA across seven distinct cognitive dimensions.}

    \label{fig:pipeline}
\end{figure*}
\section{ChinaHeritaQA: Dataset Construction}
\label{sec:data}
Constructing a benchmark for cultural heritage requires balancing visual diversity with historical rigor. We adopt a structured, multi-stage pipeline: (1) \textbf{Ontology Construction} based on UNESCO standards; (2) \textbf{In-the-Wild Image Collection} from social media; (3) \textbf{Attribute Decomposition} based on LLM; (4) \textbf{Question Formulation} based on fine-grained meta-data; and (5) \textbf{Human Verification}. Figure \ref{fig:pipeline} shows an overview of our pipeline.

% ----------------------------------------------------
% 3.1 Ontology
% ----------------------------------------------------
\subsection{Heritage Ontology and Knowledge Schema}
To ensure a comprehensive coverage of Chinese culture, we ground our dataset in the \textbf{UNESCO World Heritage List}\footnote{\url{https://whc.unesco.org/en/list/}}. This ontology covers 60 heritage sites, ensuring diversity across dynasties (from Neolithic to Qing) and functions (political, religious, residential). For each site, we collect bilingual (Chinese and English) raw textual descriptions from online heritage-related sources, including long-form encyclopedic introductions and UNESCO selection-criteria texts.

% ----------------------------------------------------
% 3.2 Image Collection
% ----------------------------------------------------
\subsection{In-the-Wild Image Collection}
Existing datasets often rely on canonical, encyclopedia-style images. To bridge the gap between benchmarks and real-world applications, we aim to capture heritage sites as they appear in ``lived experiences.''

\paragraph{Source and Crawling.} We utilized \textbf{Sina Weibo}\footnote{\url{https://weibo.com/}}, one of China's largest social media platforms, as our data source. Using specific entity names from our ontology as queries, we collected over 50,000 raw images. This approach enables us to capture heritage sites under diverse lighting conditions, viewing angles, and degrees of occlusion, mirroring the visual complexity encountered in real-world scenarios.

\paragraph{De-noising Pipeline.} Social media data is inherently noisy. We applied a rigorous filtering protocol:
\begin{enumerate}[leftmargin=*,itemsep=1pt, topsep=2pt]
    \item[1)] \textbf{Visual Quality Filter:} We removed images with low resolution ($<512px$), severe blurriness, or excessive text overlays. To ensure visual clarity, specific social media tags or watermarks on the figures were removed using automated inpainting techniques.
    \item[2)] \textbf{Privacy and Relevance Filter (CLIP + Human):} We first used CLIP \cite{pmlr-v139-radford21a} to filter out semantically irrelevant images (e.g., tickets, maps; see Appendix \ref{app:clip_negative_filter} for a detailed list of filtered categories). Subsequently, trained annotators manually discarded images containing: (a) Selfies or portraits dominating the frame; (b) Close-ups of non-heritage objects (e.g., souvenir shops, food); (c) Artistically distorted photos that lose realistic features; and (d) Privacy Preservation: Any pictures containing identifiable human faces were strictly excluded.
\end{enumerate}

After applying rigorous image quality filters and privacy screening, the dataset encompasses 51 UNESCO World Heritage sites in China with a total of 2,279 high-quality images.

% ----------------------------------------------------
% 3.4 Attribute Decomposition
% ----------------------------------------------------
\subsection{Attribute Decomposition}
Given that the collected raw texts are typically lengthy and unstructured, we leverage GPT-4o \cite{openai2024gpt4ocard} to refine them into a structured attribute schema. This process transforms unstructured descriptions into unified, site-level representations, laying a direct foundation for subsequent question generation. Specifically, we develop a comprehensive knowledge schema for each heritage site, comprising the following core dimensions:

\begin{itemize}[leftmargin=*,itemsep=1pt, topsep=2pt]
    \item \textbf{Basic Metadata:} Includes bilingual names (Chinese and English), heritage categories (cultural, natural, or mixed), and geographical locations.
    \item \textbf{Historical Background:} Specifies the associated dynasties or historical periods, along with the background of their construction.
    \item \textbf{Descriptive Knowledge:} Integrates encyclopedic summaries from Wikipedia\footnote{\url{https://www.wikipedia.org/}.} for general overviews, and incorporates official Selection Criteria texts from the UNESCO archives to elucidate the specific historical or artistic values that justify each site's status.
\end{itemize}

Finally, to ensure high data quality and accuracy, trained annotators conduct manual verification of the extracted information, retaining only the knowledge entries that reach a unanimous consensus.

% ----------------------------------------------------
% 3.5 Question Formulation & Types
% ----------------------------------------------------
\subsection{Question Formulation and Taxonomy}
Based on the curated knowledge schema, we designed \textbf{ChinaHeritaQA} to assess models across varying levels of cognitive demand. We define two task formats: \textbf{Single-Image VQA (SVQA)}, where the model answers based on one image, and \textbf{Multi-Image VQA (MVQA)}, where the model selects the correct image from a set.

All questions are presented in a multiple-choice format with five options: one correct answer and four carefully curated distractors. Specifically, the distractors are generated based on a structured sampling strategy, including a same-type site, a same-province site, another Chinese site, and a non-Chinese site. The inclusion of the non-Chinese heritage site (e.g., Western architecture) serves as a specific cross-cultural distractor, designed to test the model's resistance to hallucination and cultural confusion.

We categorized the questions into 7 distinct types:

\begin{itemize}[leftmargin=*,itemsep=1pt, topsep=3pt]

\item {\small
\textbf{Type 1 (SVQA): Identity Recognition.}

The most fundamental task requiring the model to identify the specific heritage site name given visual input.

\textit{Example:} 
\begin{CJK}{UTF8}{gbsn}
图片中展示的是以下哪处文化或自然遗产地?
(Which of the following cultural or natural heritage sites is depicted in this image?)
\end{CJK}
}

\item {\small
\textbf{Type 2 (MVQA): Visual Grounding.}

This task inverts Type 1. Given a heritage site name, the model must select the correct corresponding image from a set of candidates.

\textit{Example:} 
\begin{CJK}{UTF8}{gbsn}
以下哪个图片可能是在武当山古建筑群拍摄的?
(Which of the following images was likely taken at the Ancient Building Complex in the Wudang Mountains?)
\end{CJK}
}

\item {\small
\textbf{Type 3 (SVQA): Description Matching.}

This tests general understanding. Given an image, the model must select the correct encyclopedic summary (derived from Wikipedia meta-data).

\textit{Example:} 
\begin{CJK}{UTF8}{gbsn}
关于该图片简要介绍正确的是?
(Which brief introduction regarding this picture is correct?)
\end{CJK}
}

\item {\small
\textbf{Type 4 (SVQA): Historical Periodization.}

The model must identify the specific dynasty or era when the architecture in the image was constructed. Distractors include dynasties from different eras and non-Chinese historical periods.

\textit{Example:} 
\begin{CJK}{UTF8}{gbsn}
该图片中的建筑群可能建于哪个朝代?
(In which dynasty might the building complex in this picture have been built?)
\end{CJK}
}

\item {\small
\textbf{Type 5 (SVQA): Historical Contextualization.}

An advanced reasoning task beyond simple dynasty naming. It asks for the specific historical background or events associated with the site's construction, retrieved from Wikipedia meta-data.

\textit{Example:}
\begin{CJK}{UTF8}{gbsn}
关于该图片历史背景介绍正确的是?
(Which description of the historical background of this picture is correct?)
\end{CJK}
}

\item {\small
\textbf{Type 6 (SVQA): Functional Analysis.}

The model must infer the primary function of the site (e.g., religious worship, military defense, royal residence) based on visual cues and cultural knowledge.

\textit{Example:} 
\begin{CJK}{UTF8}{gbsn}
关于该图片主要的功能介绍正确的是?
(Which description of the main function of this picture is correct?)
\end{CJK}
}

\item {\small
\textbf{Type 7 (SVQA): Architectural Analysis.}

This probes fine-grained visual reasoning regarding architectural style, structural components, or usage specific to the building's design.

\textit{Example:} 
\begin{CJK}{UTF8}{gbsn}
关于该图片建筑用途介绍正确的是?
(Which description of the architectural usage of this picture is correct?)
\end{CJK}
}

\end{itemize}

Detailed examples with pictures are presented in Appendix \ref{app:question_examples}.

% ----------------------------------------------------
% 3.6 Verification
% ----------------------------------------------------
\subsection{Human Verification}
To ensure the ``Gold Standard'' quality of our benchmark, we implemented a strict verification phase. A separate group of annotators reviewed each $(Image, Question, Answer)$ triplet to perform:
(1) \textbf{Solvability Check:} Ensuring the question can be answered using the visual information and cultural knowledge; and
(2) \textbf{Fact Verification:} Cross-referencing answers with UNESCO dossiers.
Ambiguous or grammatically incorrect items were flagged and removed.

\section{Benchmark Characteristics}
In this section, we conduct an in-depth statistical and feature analysis of the ChinaHeritaQA dataset across four dimensions: overall scale, heritage attributes, temporal span, and geographical diversity. 
% ----------------------------------------------------
% 4.1 Overall  
% ----------------------------------------------------
\subsection{Overall Statistics}
Table \ref{tab:overall_statistics} shown that ChinaHeritaQA contains 14,133 bilingual multiple-choice QA pairs built from 2,279 filtered in-the-wild images, covering 51 UNESCO-defined Chinese World Heritage sites. The benchmark supports both Single-Image VQA and Multi-Image VQA and includes seven question types. Each question is presented with five options.
\begin{table}[htbp]
\centering
\small
\renewcommand{\arraystretch}{1.00}
\begin{tabular}{l r}
\toprule
\textbf{Statistic} & \textbf{Value} \\
\midrule
Heritage sites & 51 \\
Raw images & 50,000+ \\
Final images & 2,279 \\
QA pairs & 14133 \\
Languages & Chinese / English \\
Task formats & SVQA / MVQA \\
Question types & 7 \\
Options per question & 5 \\
\bottomrule
\end{tabular}
\caption{Overall statistics of ChinaHeritaQA.}
\label{tab:overall_statistics}
\end{table}

\subsection{Question Type Coverage}

Table~\ref{tab:qtype_statistics} summarizes the question-type coverage and bilingual input length of ChinaHeritaQA. The dataset contains 14,133 multiple-choice QA pairs across seven question types. Q1, Q2, Q3, and Q6 each contain 2,279 QA pairs, while Q5, Q4, and Q7 contain 1,989, 1,658, and 1,370 QA pairs, respectively. This distribution provides broad coverage for both basic visual recognition and culturally grounded reasoning.
\begin{table}[htbp]
\centering
\scriptsize
\setlength{\tabcolsep}{4pt}
\renewcommand{\arraystretch}{1.00}
\begin{tabular}{@{}llrrrr@{}}
\toprule
Type & Question Type & \#QA & Ratio & Avg. CN & Avg. EN \\
\midrule
Q1 & Identity & 2,279 & 16.13\% & 203.8 & 159.7 \\
Q2 & Grounding & 2,279 & 16.13\% & 206.8 & 162.6 \\
Q3 & Description & 2,279 & 16.13\% & 458.6 & 290.9 \\
Q4 & Periodization & 1,658 & 11.73\% & 176.5 & 135.4 \\
Q5 & Contextualization & 1,989 & 14.07\% & 239.6 & 192.4 \\
Q6 & Function & 2,279 & 16.13\% & 216.2 & 152.7 \\
Q7 & Architecture & 1,370 & 9.69\% & 222.2 & 155.4 \\
\midrule
Total & -- & 14,133 & 100.00\% & 251.0 & 181.5 \\
\bottomrule
\end{tabular}
\caption{Question-type statistics of ChinaHeritaQA. Avg. CN and Avg. EN denote the average token lengths of the question stem and five answer options, excluding system prompts and evaluation instructions.}
\label{tab:qtype_statistics}
\end{table}

\subsection{Chronological Distribution}
\label{sec:chronological}

\begin{figure}[htbp]
    \centering
    \includegraphics[width=0.5\textwidth]{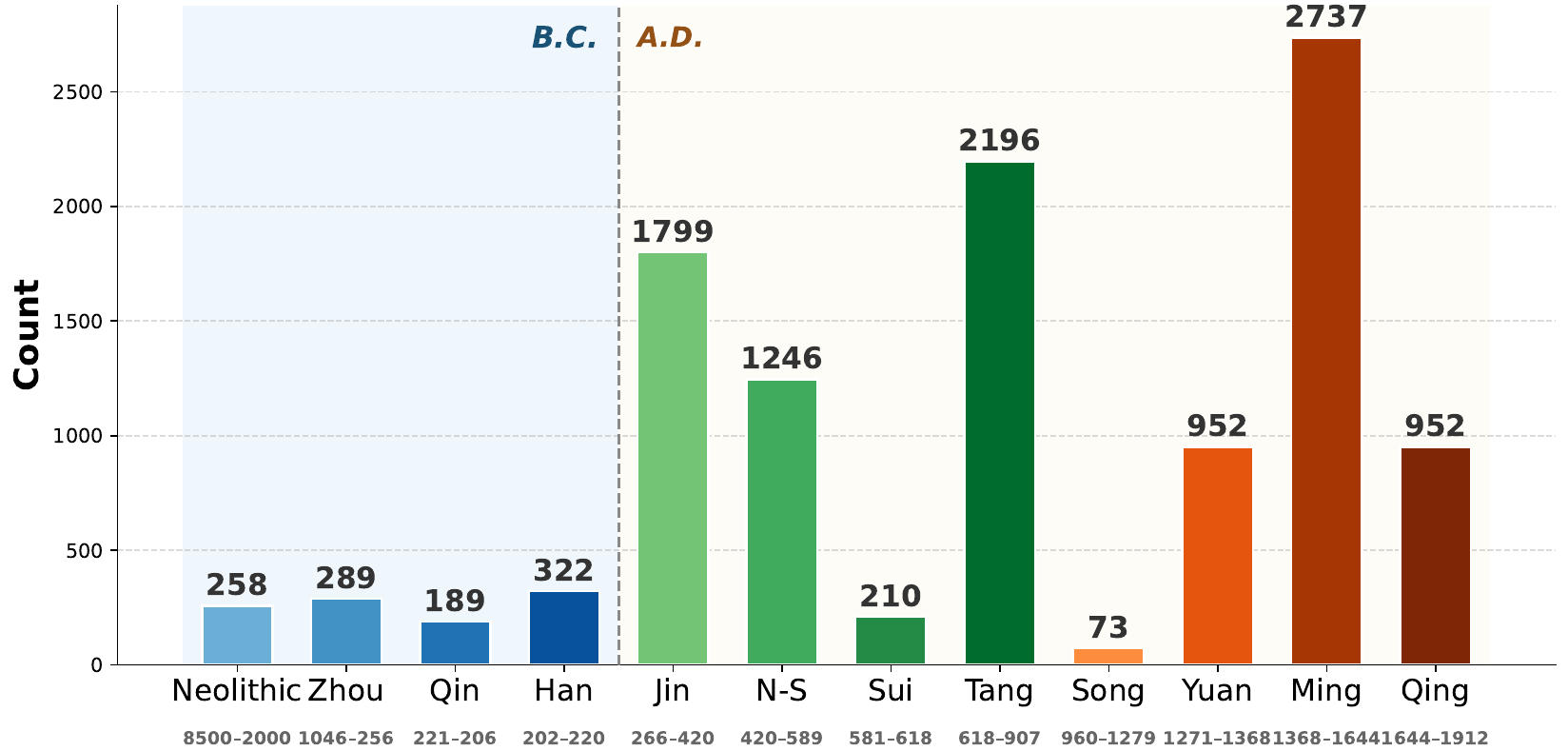}
    \caption{The chronological distribution of QA pairs in ChinaHeritaQA.}
    \label{fig:dynast_dis}
\end{figure}

As shown in Figure \ref{fig:dynast_dis}, the chronological distribution of \text{ChinaHeritaQA} exhibits dramatic fluctuations rather than a smooth curve, directly mirroring the historical survivorship bias of cultural heritages. The time period is based on Wikipedia.\footnote{\url{https://en.wikipedia.org/wiki/History_of_China}}

The Ming, Tang, and Jin dynasties dominate the timeline. This abundance is primarily driven by the high survival rate of relatively recent brick-and-wood structures (Ming) and the extensive preservation of stone grottoes and murals (Tang, Jin).

Conversely, eras such as the Song, Qin, and Sui dynasties present severe data scarcity due to the vulnerability and poor preservation of early wooden architectures.

An additional 2,289 instances belong to Natural or Mixed heritages (e.g., South China Karst, Mount Emei Scenic Area, including Leshan Giant Buddha Scenic Area) lacking a specific dynastic timeline.

\subsection{Geographical Distribution}
\label{sec:geographical}

\begin{figure}[htbp]
    \centering
    \includegraphics[width=0.9\linewidth]{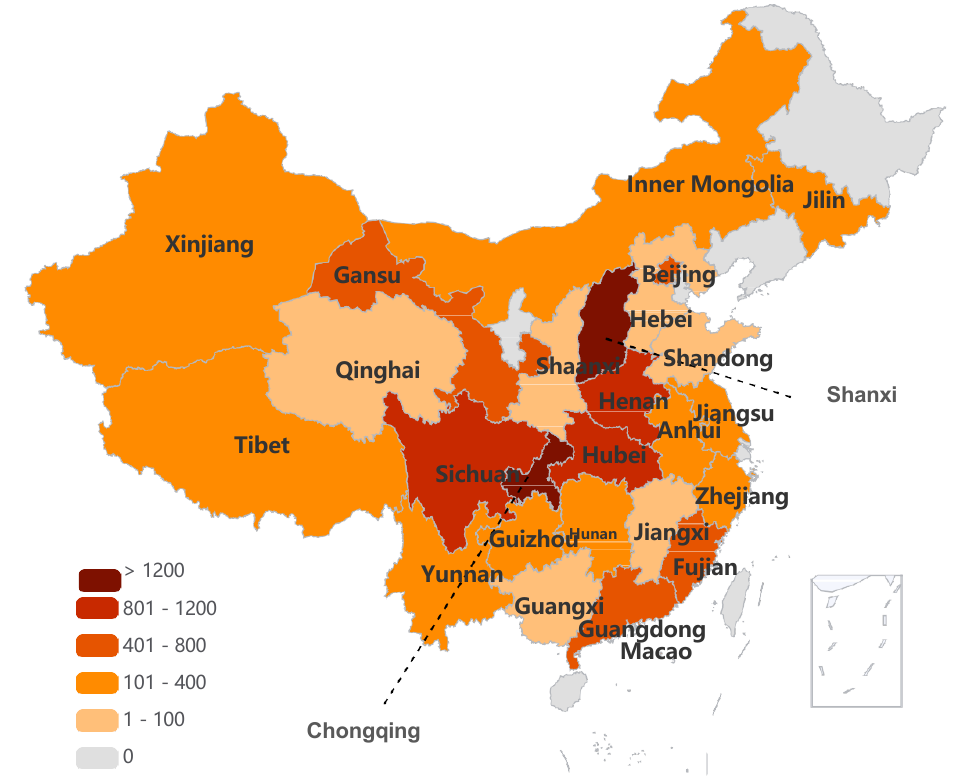}
    \caption{Geographical heatmap of QA pairs.}
    \label{fig:geo_map}
\end{figure}

As shown in Figure~\ref{fig:geo_map}, the QA pairs are unevenly distributed across China's provinces. 

The data is mostly concentrated in central and southwestern China, particularly in Shanxi (1,812 pairs) and Chongqing (1,565 pairs). Shanxi has many well-preserved ancient wooden buildings and grottoes, while Chongqing features unique landscapes and cultural sites. Both generate high user interest and abundant photos on social media. In contrast, many regions have very little data. Provinces like Qinghai, Jiangxi, and Shandong each have fewer than 100 instances. Some areas (grey zones) have no data due to a lack of relevant UNESCO sites or our strict filtering. This imbalance provides a good test for the models' ability to recognize diverse regional features.

\begin{table*}[htbp]
\centering
\scriptsize
\setlength{\tabcolsep}{2.5pt}
\renewcommand{\arraystretch}{1.0}

\resizebox{\textwidth}{!}{%
\begin{tabular}{@{}l*{16}{c}@{}}
\toprule
\multirow{2}{*}{} 
& \multicolumn{2}{c}{Q1} 
& \multicolumn{2}{c}{Q2} 
& \multicolumn{2}{c}{Q3} 
& \multicolumn{2}{c}{Q4} 
& \multicolumn{2}{c}{Q5} 
& \multicolumn{2}{c}{Q6} 
& \multicolumn{2}{c}{Q7} 
& \multicolumn{2}{c}{Avg} \\
\cmidrule(lr){2-3} 
\cmidrule(lr){4-5} 
\cmidrule(lr){6-7} 
\cmidrule(lr){8-9} 
\cmidrule(lr){10-11} 
\cmidrule(lr){12-13} 
\cmidrule(lr){14-15} 
\cmidrule(lr){16-17}
& Acc & F1 & Acc & F1 & Acc & F1 & Acc & F1 
& Acc & F1 & Acc & F1 & Acc & F1 & Acc & F1 \\
\midrule
CogVLM2-19B            & 79.46 & 79.45 & –     & –     & 53.14 & 48.40 & 49.40 & 47.26 & 50.58 & 49.79 & 54.01 & 53.40 & 51.02 & 50.52 & 57.17 & 56.43 \\
Deepseek-vl2-small     & 85.28 & 85.30 & 74.75 & 74.62 & 65.47 & 65.15 & 61.22 & 61.22 & 64.48 & 64.39 & 63.32 & 63.31 & 65.22 & 65.23 & 69.15 & 69.14 \\
GLM-4.6V-Flash         & 80.78 & 80.85 & 82.51 & 82.33 & 70.18 & 69.41 & 54.31 & 54.00 & 68.10 & 67.88 & 64.70 & 64.47 & 68.54 & 68.21 & 70.68 & 70.42 \\
InternVL2.5-8B         & 89.10 & 89.08 & 80.78 & 80.65 & 72.27 & 72.20 & 58.38 & 58.40 & 72.40 & 72.37 & 67.22 & 67.21 & 67.55 & 67.54 & 73.47 & 73.47 \\
Qwen2.5-VL-7B-Instruct & 95.06 & 95.04 & 90.85 & 90.85 & 75.89 & 75.84 & 63.09 & 63.05 & 78.33 & 78.31 & 74.09 & 74.05 & 78.65 & 78.64 & 80.21 & 80.21 \\
Qwen3-VL-8B-Instruct   & 95.09 & 95.08 & 92.96 & 93.15 & 80.72 & 80.59 & 64.42 & 64.43 & 78.38 & 78.37 & 74.16 & 74.12 & 78.65 & 78.66 & 81.51 & 81.54 \\
\midrule
Human & 76.00 & 75.50 & 84.00 & 82.00  & 61.90 & 61.90 & 44.70 & 43.20 & 64.70 & 64.30 & 66.70 & 65.60 & 61.30 & 60.20 & 67.30 & 67.20 \\
\bottomrule
\end{tabular}%
}
\caption{Performance of evaluated VLMs on ChinaHeritaQA across seven question types. We report accuracy and macro-F1 for each question type and the averaged results over all supported tasks. “–” denotes unavailable results, as CogVLM2-19B does not support the multi-image in Q2 setting.}
\label{tab:overall_results}
\end{table*}
\section{Experiments and Results}

\subsection{Experimental Setup}
\paragraph{Models.} We experiment with open-weighted VLMs including CogVLM2-19B \cite{hong2024cogvlm2visuallanguagemodels}, Deepseek-vl2-small \cite{wu2024deepseekvl2mixtureofexpertsvisionlanguagemodels}, InternVL2.5-8B \cite{chen2024internvlscalingvisionfoundation}, GLM-4.6V-Flash \cite{vteam2026glm45vglm41vthinkingversatilemultimodal}, Qwen2.5-VL-7B-Instruct \cite{bai2025qwen25vltechnicalreport}, Qwen3-VL-8B-Instruct \cite{yang2025qwen3technicalreport}. Based on our VQA framework, we prompt these models using five task-specific Chinese instructions that clearly define task roles, descriptions, and output requirements. Detailed prompts are shown in Appendix \ref{app:prompts}. Additional setups are shown in Appendix \ref{app:Experiment}.

\paragraph{Human Performance Evaluation.}
We conduct a human baseline via stratified sampling. Native Chinese speakers answered a representative subset of 350 QA pairs (50 from each of the 7 cognitive dimensions), providing a reference to measure the true gap between VLMs and human cultural reasoning. Details are shown in Appendix \ref{app:human}.

\subsection{Overall Performance}

Table~\ref{tab:overall_results} summarizes the results of all evaluated VLMs on ChinaHeritaQA. 
Overall, the Qwen-series models achieve the strongest performance. 
Qwen3-VL-8B-Instruct obtains the best average accuracy and macro-F1, reaching 81.51\% and 81.54\%, respectively, followed by Qwen2.5-VL-7B-Instruct with 80.21\% accuracy and 80.21\% macro-F1. 

\textbf{VLM vs. Human.} Compared with the human performance, top-performing VLMs show clear advantages on several tasks. 
For example, Qwen3-VL-8B-Instruct achieves 95.09\% accuracy on Q1 and 92.96\% on Q2, while the human baseline reaches 76.00\% and 84.00\% on the same two tasks. 
This suggests that strong VLMs can effectively leverage large-scale pretraining knowledge to recognize visually salient heritage sites and associate them with known cultural entities. 

\begin{figure}[htbp]
    \centering
    \includegraphics[width=0.95\linewidth]{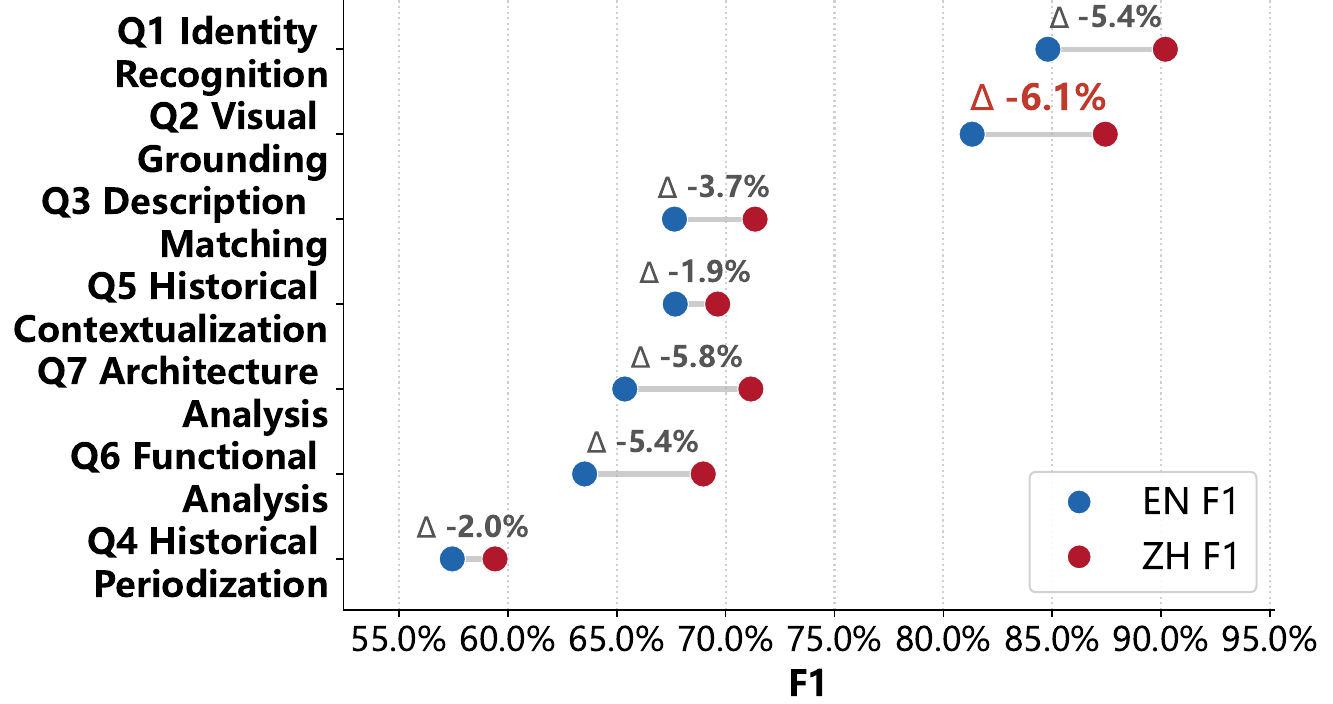}
    \caption{F1 comparison across question types in Chinese and English. Blue and red markers denote English and Chinese F1 scores, respectively, with horizontal gaps indicating the performance difference between the two language settings.}
    \label{fig:cross_linguistic}
\end{figure}

\textbf{ZH vs EN.} Models consistently perform better in Chinese than English (Figure~\ref{fig:cross_linguistic}), with the largest drop on Q2 (-6.1\% F1). Large gaps also appear in Q7, Q1, and Q6 (tasks requiring precise alignment between cultural terms, visual evidence, and heritage concepts), thus more sensitive to translation. In contrast, Q5, Q4, and Q3 show smaller gaps, as they rely more on encyclopedic knowledge than culturally specific terminology. This indicates that cross-lingual degradation is most severe when visual grounding depends on native cultural names or architectural terminology.

\textbf{Recognition vs. Visual Grounding.} Q1 (single-image recognition) and Q2 (multi-image grounding) both evaluate heritage-site recognition but under different formats. Models achieve 87.46\% accuracy on Q1 and 84.37\% on Q2, indicating that multi-image grounding introduces additional difficulty. This shows that within recognition tasks, answer format significantly affects model behavior: single-image recognition tests image-to-entity association, while multi-image recognition further tests cross-image discrimination.

\section{Further Analysis}
\label{sec:Analysis}

We further analyze VLM performance across question types, 
dynasties, 
and geographical regions. We also provide an error analysis in the end.

\textbf{Strong visual recognition does not imply deep cultural understanding.} 

\begin{figure}[htbp]
    \centering
    \includegraphics[width=\linewidth]{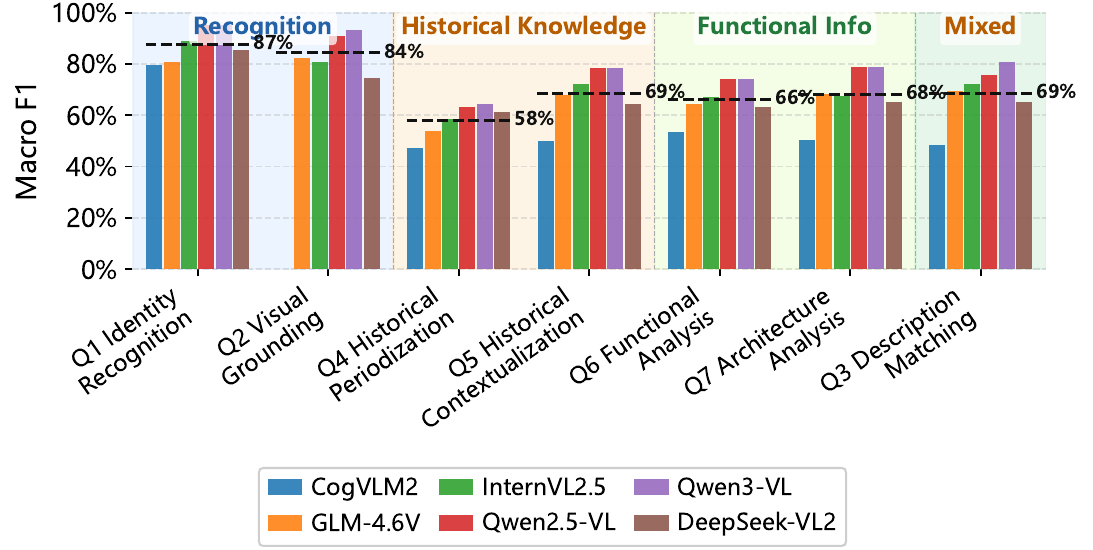}
    \caption{Macro-F1 scores for seven question types, grouped into four broader capability categories.}
    \label{fig:qtype_model_line}
\end{figure}

Figure \ref{fig:qtype_model_line} further classifies the questions into four broader capability categories and shows the performance differences. 
Model performance varies significantly across question types. VLMs excel at recognition tasks (87.46\% on Q1, 84.37\% on Q2) but show substantial gaps on reasoning-oriented categories. This suggests that current VLMs can associate visually distinctive heritage images with entities but struggle to explain the historical, functional, and architectural meanings behind them.

\textbf{Historical periodization is a common challenge.} Almost all models reach lowest performance on Q4 (historical periodization), with the best model achieving only 64\%. In contrast, performance recovers on Q5 (historical contextualization), indicating that models are better at selecting from semantically rich options than directly inferring periods from visual evidence. The core weakness is not insufficient historical knowledge, but the lack of effective image-to-period grounding.

\begin{figure}[htbp]
    \centering
    \includegraphics[width=\linewidth]{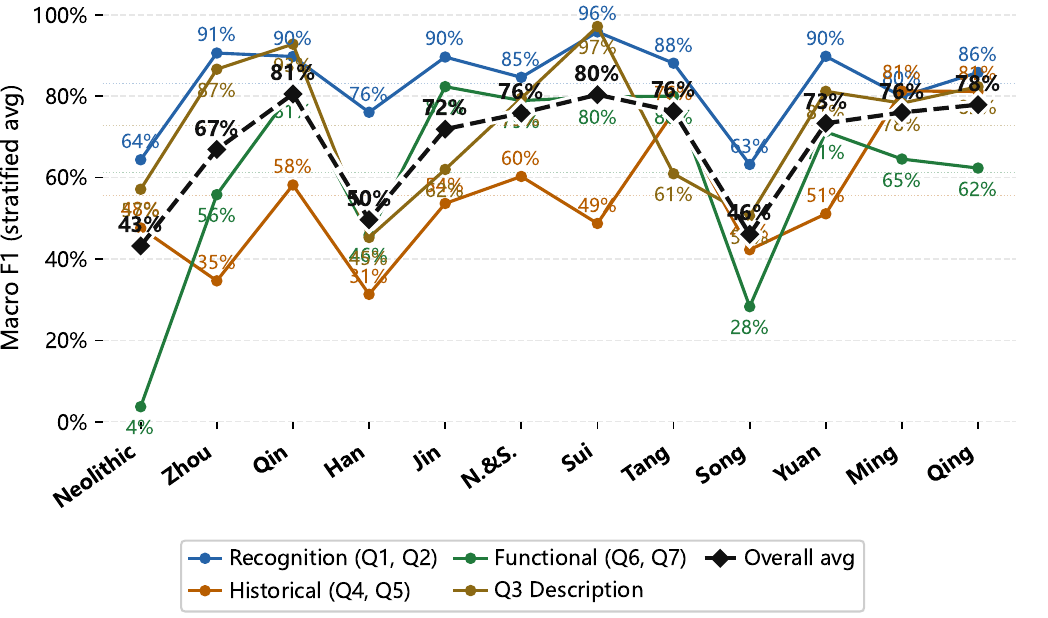}
        \caption{Mean Macro-F1 across dynasties, grouped into four capability types.}
    \label{fig:dynasty_category_line}
\end{figure}

\textbf{VLMs exhibit dynasty-level temporal grounding bias.} Ming, Qing, and Sui achieve stronger performance, while Neolithic, Han, and Song are weaker. This reflects uneven distribution of heritage knowledge in pretraining data. Historical judgment fails especially for Zhou, Han, Song, and Neolithic, where models identify sites but fail to connect them to correct periods. Functional judgment is weakest for Neolithic. Overall, two key boundaries emerge: image-to-history grounding and image-to-function grounding (Figure \ref{fig:dynasty_category_line}).

\begin{figure}[htbp]
    \centering
    \includegraphics[width=\linewidth]{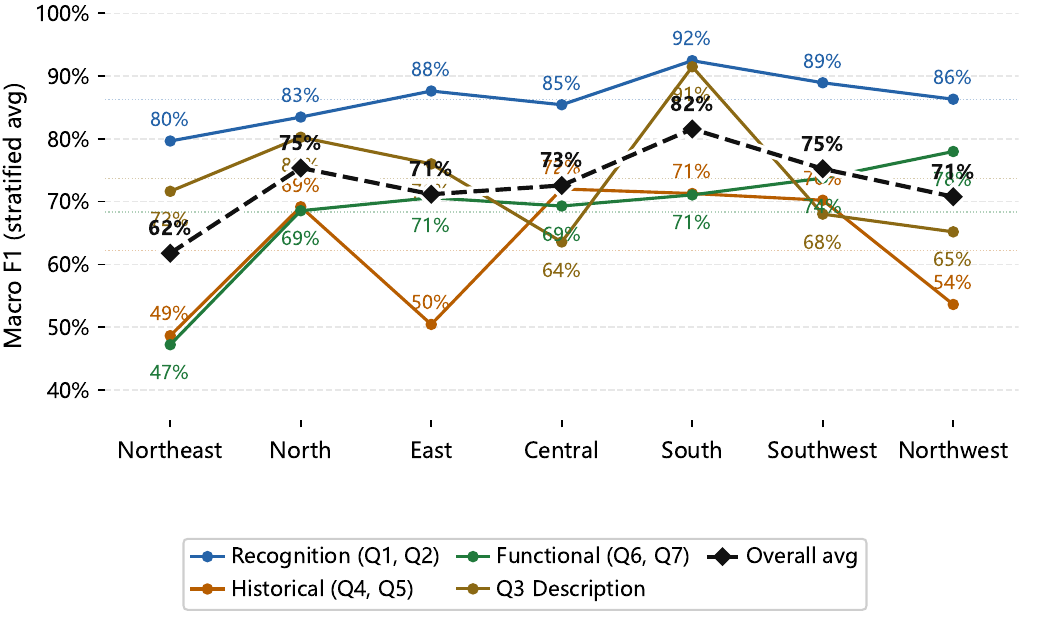}
    \caption{Mean Macro-F1 across seven macro-regions, grouped into four capability categories.}
    \label{fig:region_category_line}
\end{figure}

\textbf{VLMs show region-level cultural grounding bias.} 
Figure \ref{fig:region_category_line} shows the performance in seven macro-regions. South China shows strong performance across recognition and description matching. East China reveals a dispersed profile: high recognition but low historical grounding, suggesting models identify eastern sites but fail to contextualize them historically. Northeast and Northwest China expose weaknesses in historical and functional interpretation, indicating incomplete encoding of regional heritage patterns.

\begin{figure}[htbp]
    \centering
    \includegraphics[width=0.95\linewidth]{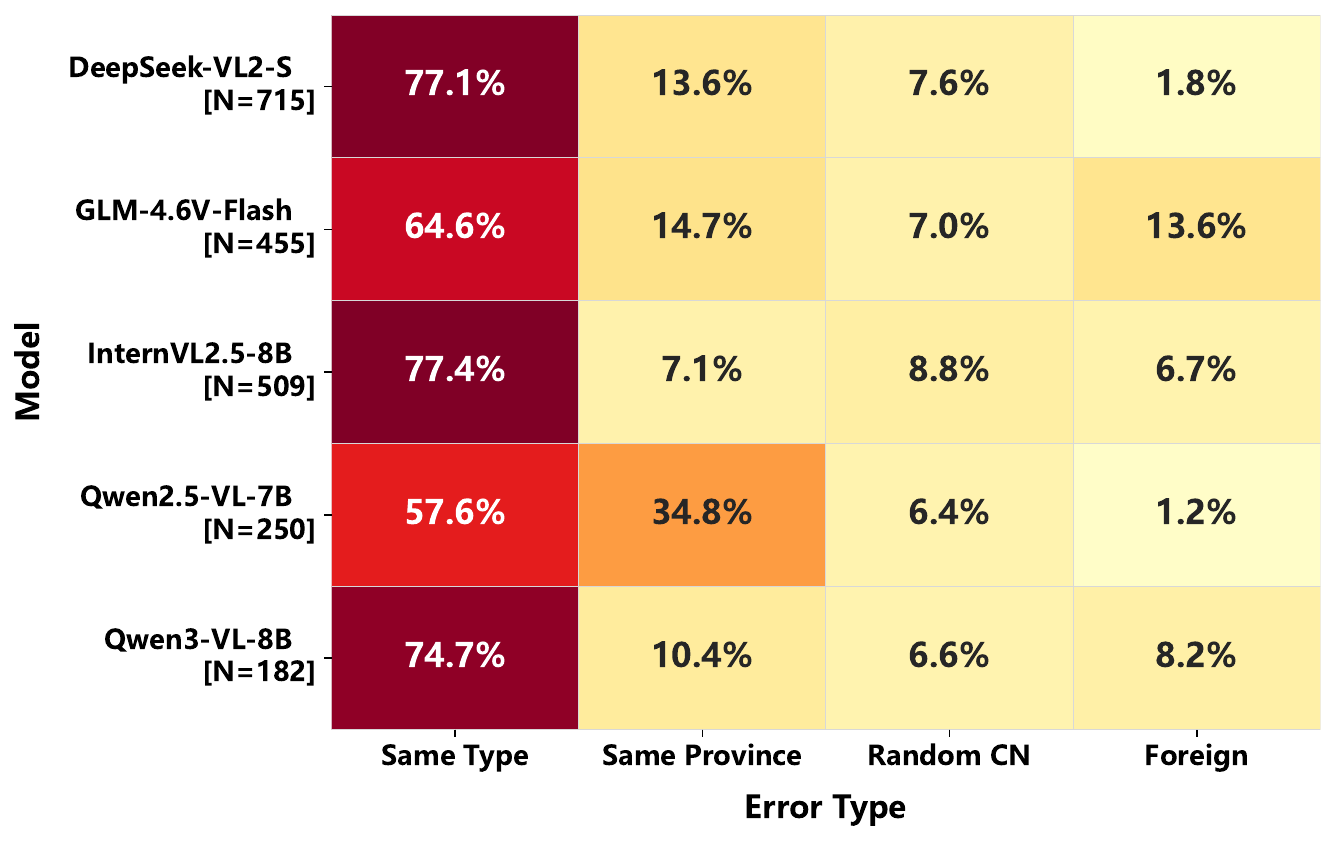}
    \caption{Distribution of wrong-answer types for Q2 across VLMs.}
    \label{fig:q2_error_heatmap_filtered}
\end{figure}

\begin{figure}[htbp]
    \centering
    \includegraphics[width=\linewidth]{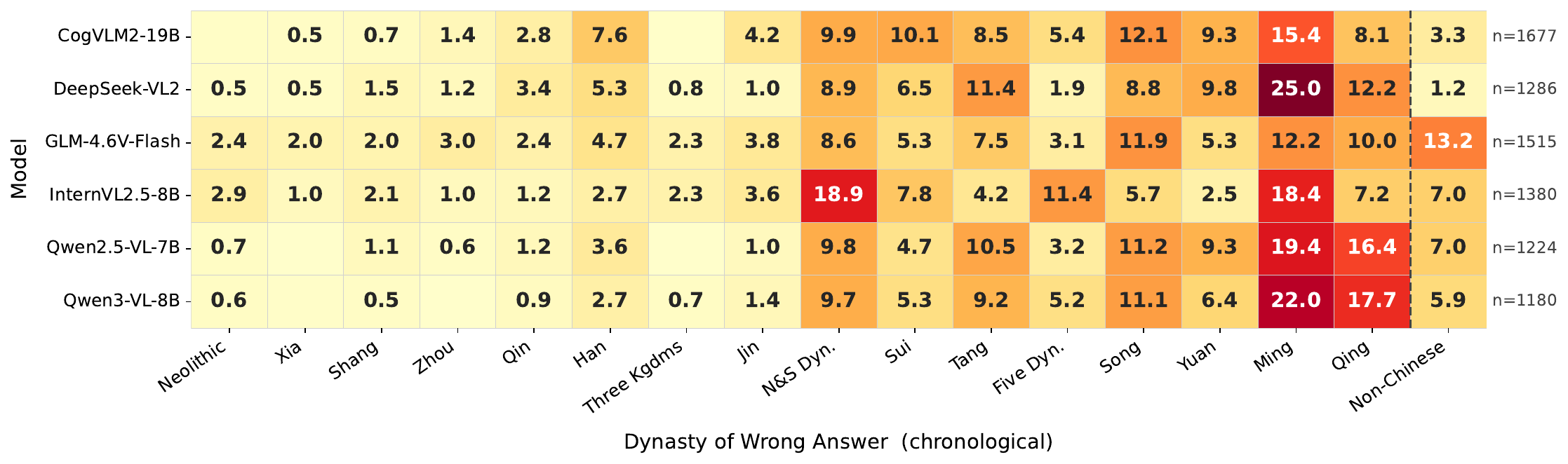}
    \caption{Distribution of wrong-answer dynasties for Q4 across VLMs.}
    \label{fig:q4_dynasty_error_heatmap}
\end{figure}

\paragraph{Error Analysis.} To understand why models fail on specific questions, we conduct an additional error analysis on Q2 and Q4, the two tasks that represent VLM performances in visual grounding (Q2) and temporal reasoning (Q4) (Figures \ref{fig:q2_error_heatmap_filtered}--\ref{fig:q4_dynasty_error_heatmap}).
 
For Q2, most errors concentrate on same-type distractors (57.6\%--77.4\%), indicating that models recognize heritage categories but fail at fine-grained site discrimination. Appendix \ref{app:error} showcases an example for this case. The bottleneck is not cross-cultural confusion but visual grounding among visually similar Chinese sites.
 
For Q4, errors cluster around historically salient dynasties (Ming, Qing, Song, Tang), rather than random guessing. This suggests that when visual evidence is insufficient, models default to familiar historical priors rather than architectural details. Both patterns reveal a common theme: models lack robust mappings between visual heritage features and specific cultural entities or historical periods, even when the underlying knowledge exists.
 
\section{Conclusion}
We introduced ChinaHeritaQA, a bilingual multimodal benchmark with 2,279 images and 14,133 QA pairs covering Chinese UNESCO World Heritage sites. While VLMs outperform humans on average, they struggle with culturally grounded reasoning tasks. Performance varies substantially by dynasty and region, revealing that visual retrieval does not straightforwardly extend to cultural and historical understanding. We release the dataset to support future research on culturally aware multimodal learning.

\section*{Limitations}

ChinaHeritaQA covers 51 of China's 60 UNESCO World Heritage sites, with substantial geographic imbalance: Shanxi and Chongqing account for nearly 40\% of QA pairs, while several provinces contribute fewer than 100 instances. This is mainly due to the fact that the extracted pictures in Sina Weibo only covered these 51 sites. This imbalance may conflate data availability with cultural reasoning ability. Similarly, chronological distribution exhibits pronounced survivorship bias: Ming, Tang, and Jin dynasties comprise over 60\% of instances, while Song and Qin have fewer than 250 pairs each. As a result, model performance gaps on underrepresented dynasties may reflect data scarcity rather than inherent reasoning limitations.
 
Our human baseline comprises three college-educated native speakers, not heritage experts. Inter-evaluator agreement on historical periodization (Q4) reached only 16.0\% ($\kappa = 0.247$, see Appendix \ref{app:human}), indicating genuine ambiguity even for informed humans. The five-option multiple-choice format may also introduce artifacts: carefully curated distractors could inflate difficulty for some tasks while reducing it for others compared to open-ended evaluation.
 
All images are sourced from Sina Weibo, introducing selection bias toward visually distinctive, photogenic sites. Bilingual evaluation requires translation from Chinese originals; performance gaps between languages may reflect translation quality rather than linguistic-cultural differences. We evaluate only open-weight models; findings may not generalize to closed-weight systems with different training data. Finally, ChinaHeritaQA is grounded in the UNESCO heritage framework, which may not capture locally-defined or contested heritage narratives outside this international regime.

\section*{Ethical Considerations}
We address potential ethical considerations arising from the construction and use of ChinaHeritaQA.
 
\paragraph{Dataset Source and Intellectual Property.} ChinaHeritaQA draws images from Sina Weibo, China's largest social media platform, where heritage site photos are shared publicly. All images are collected under Sina Weibo's terms of service and are used solely for research purposes. Textual descriptions of heritage sites are sourced from publicly available Wikipedia entries and UNESCO official selection criteria documents. We ensure that this dataset is intended exclusively for research and educational purposes and should not be used for commercial applications. The dataset construction adheres strictly to the intellectual property requirements of the source materials and respects the privacy and attribution rights of original content creators.
 
\paragraph{Image Privacy and Content Filtering.} During the image collection and de-noising pipeline, we implemented rigorous privacy protections. Specifically, any images containing identifiable human faces were automatically excluded to protect the privacy of social media users. Additionally, we removed images with excessive personal information or metadata that could compromise individual privacy. This filtering was performed both through automated CLIP-based filtering and manual human review to ensure comprehensive privacy protection.
 
\paragraph{Data Annotation and Human Evaluation.} Before annotation, all human annotators were fully informed about the task objectives, data usage, and ethical guidelines. All annotators and human evaluators were project partners who voluntarily contributed to the dataset, and are native Chinese speakers with college education and cultural heritage knowledge.
 
\paragraph{Potential Biases and Limitations.} The dataset reflects the visual perspectives of social media users, which may skew toward photogenic, iconic vantage points and exclude certain heritage aspects. Furthermore, the bilingual nature of the dataset means English translations of Chinese heritage terminology may not fully capture cultural nuance. We acknowledge these limitations and encourage users to consider them when interpreting results. The dataset does not contain personally identifiable information beyond the inherent metadata in public social media images.
 
\paragraph{Use of AI Tools.} This work employed GPT-4o for attribute decomposition and question generation (Section \ref{sec:data}), with outputs subsequently verified by human annotators to ensure accuracy and cultural appropriateness. The authors also acknowledge the use of Claude AI for manuscript refinement, including structure organization and clarity enhancement. All uses of AI tools were supplementary to human judgment and subject to human verification.

\section*{Acknowledgments}
DH was supported by Independent Research Fund Denmark under grant ID 10.46540/5334-00088B.

% Bibliography entries for the entire Anthology, followed by custom entries
%\bibliography{anthology,custom}
% Custom bibliography entries only
\bibliography{custom}

\appendix
\section{CLIP-based Forced Negative Filtering}
\label{app:clip_negative_filter}

To reduce obvious noise in social-media images, we use CLIP \cite{pmlr-v139-radford21a} as a conservative negative semantic filter. The goal of this step is not to identify whether an image belongs to a specific heritage site, but only to remove images that are highly likely to be irrelevant to heritage-site visual understanding. Specifically, each crawled image is compared against a predefined set of negative prompts describing common non-heritage tourism content. Images are automatically removed only when their maximum similarity to a negative category exceeds the category-specific threshold. All remaining images are retained for subsequent human relevance and privacy verification.

Given an image $I$ and a set of negative text prompts $\mathcal{P}_c$ for category $c$, we compute the CLIP cosine similarity between the image embedding and each text embedding. The category score is defined as:

\[
s_c(I) = \max_{p \in \mathcal{P}_c} \cos \left( f_I(I), f_T(p) \right),
\]

where $f_I$ and $f_T$ denote the CLIP image and text encoders, respectively. The image is removed if:

\[
\max_c s_c(I) \geq \tau_c,
\]

where $\tau_c$ is the forced-removal threshold for the category with the highest negative score. Otherwise, the image is retained for later human inspection.

We intentionally exclude categories such as tourists, crowds, selfies, night scenes, and architectural close-ups from the forced-removal filter, since these cases may still contain valid visual evidence of heritage sites. This design makes the CLIP module a high-precision rejection filter rather than a positive heritage-site classifier. See Table \ref{tab:clip_negative_prompts}.

\section{Prompts Used for Evaluation}
\label{app:prompts}
Figure \ref{fig:system_prompts} presents the shared system prompt used for model evaluation.  
Since the same system prompt is used across all question types, it is shown once and omitted from individual examples for readability.

\begin{figure}[htbp]
    \centering
    % ============ 第一部分：中文原版 ============
    \begin{tcolorbox}[
        colback=gray!2!white, colframe=gray!70!black, 
        title={System Prompt ( Chinese)}, 
        fonttitle=\small\bfseries, boxrule=0.5pt, sharp corners,
        left=6pt, right=6pt, top=4pt, bottom=4pt
    ]
    \footnotesize
    \begin{CJK}{UTF8}{gbsn}
        你是一个专业的文化与自然遗产识别系统。请仔细观察题目中的图片，并认真阅读选项 A、B、C、D、E 的内容，结合图片信息从中选出唯一正确答案。

    \textbf{要求：}
    \begin{itemize}[leftmargin=1.4em, itemsep=0pt, topsep=2pt]
        \item 只输出正确选项的字母（A / B / C / D / E）
        \item 不要添加任何解释、标点或其他内容
    \end{itemize}
    \end{CJK}
    \end{tcolorbox}
    
    \vspace{2mm} % 两个框之间的间距
    
    % ============ 第二部分：英文翻译 ============
    \begin{tcolorbox}[
        colback=gray!2!white, colframe=gray!70!black, 
        title={System Prompt (English)}, 
        fonttitle=\small\bfseries, boxrule=0.5pt, sharp corners,
        left=6pt, right=6pt, top=4pt, bottom=4pt
    ]
    \footnotesize
You are a professional cultural and natural heritage recognition system. Please carefully examine the image in the question and read options A, B, C, D, and E. Select the single correct answer based on the image information.

Requirements:
\begin{itemize}[leftmargin=1.4em, itemsep=0pt, topsep=2pt]
    \item Output only the letter of the correct option (A / B / C / D / E)
    \item Do not include any explanation, punctuation, or additional content
\end{itemize}
    \end{tcolorbox}
    
    \caption{System prompts for VLMs in Chinese and English.}
    \label{fig:system_prompts}
\end{figure}

\section{Human Evaluation}
\label{app:human}

To establish a human baseline for our cultural heritage dataset, we conducted a human evaluation study. We recruited a panel of three internal evaluators (project partners and co-authors) who are educated college students. All evaluators are native Chinese speakers with the necessary background knowledge to assess the dataset. 

The evaluation was hosted via an online survey platform, where participants were presented with multimodal contexts containing both visual evidence and textual multiple-choice questions. Evaluators were strictly instructed to rely on the provided visual clues and their own knowledge without using external search engines. The complete instructions provided to the evaluators are shown in Figure~\ref{fig:human_eval_instructions}.

The human evaluation results across different question types are summarized in Table~\ref{tab:human_answer_agreement}. Overall, the three evaluators achieved a moderate level of consensus, with an overall exact agreement of 45.4\% and a Fleiss' $\kappa$ of 0.592. The agreement levels vary significantly across categories, reflecting the varying difficulty of the cultural heritage tasks. Evaluators demonstrated high consistency in identity recognition (Q1), visual grounding (Q2), description matching (Q3), and functional analysis (Q6), where Fleiss' $\kappa$ scores exceeded 0.62. In contrast, historical periodization (Q4) yielded the lowest consensus (16.0\% exact agreement, $\kappa = 0.247$), indicating that dating cultural artifacts solely from visual clues remains highly challenging even for human evaluators. Moderate agreement was observed for historical contextualization (Q5) and architectural analysis (Q7), further confirming the dataset's multi-tiered difficulty and its viability as a rigorous benchmark.

\begin{figure}[t]
    \centering
    % ============ 第一部分：中文原版 ============
    \begin{tcolorbox}[
        colback=gray!2!white, colframe=gray!70!black, 
        title={Instructions (Original Chinese)}, 
        fonttitle=\small\bfseries, boxrule=0.5pt, sharp corners,
        left=6pt, right=6pt, top=4pt, bottom=4pt
    ]
    \footnotesize
    \begin{CJK}{UTF8}{gbsn}
    \noindent 您好！感谢您参与本次学术评测。本研究旨在评估人工智能模型对中国文化与自然遗产相关知识的理解与推理能力。问卷包含基于图片与文本的多项选择题，涵盖历史、建筑及艺术等多方面知识。\\
    \noindent \textbf{填答指南：}
    \begin{itemize}[leftmargin=1.5em, noitemsep, topsep=2pt]
        \item \textbf{仔细观察：} 请先仔细查看题目上方提供的图片证据。
        \item \textbf{客观选择：} 根据图片细节及自身知识储备，选出最准确的一项。
        \item \textbf{真实反馈：} 请勿使用搜索引擎。我们关注的是题目区分度与模型的常识对齐情况。
    \end{itemize}
    \noindent 问卷完全匿名，数据仅用于学术研究。预计用时 15 分钟。再次感谢！
    \end{CJK}
    \end{tcolorbox}
    
    \vspace{2mm} % 两个框之间的间距
    
    % ============ 第二部分：英文翻译 ============
    \begin{tcolorbox}[
        colback=gray!2!white, colframe=gray!70!black, 
        title={Instructions (English Translation)}, 
        fonttitle=\small\bfseries, boxrule=0.5pt, sharp corners,
        left=6pt, right=6pt, top=4pt, bottom=4pt
    ]
    \footnotesize
    \noindent Thank you for participating! This study benchmarks AI models on China's Cultural and Natural Heritage. The survey consists of multimodal multiple-choice questions covering history, architecture, and art.\\
    \noindent Guidelines for Participants:
    \begin{itemize}[leftmargin=1.5em, noitemsep, topsep=2pt]
        \item Examine Image: Carefully review the visual evidence for each question.
        \item Select Best Fit: Choose the most accurate option based on visual clues and your knowledge.
        \item Authentic Response: Do not use external search engines. We aim to measure dataset difficulty.
    \end{itemize}
    \noindent Participation is anonymous and for research purposes only. Estimated time: 15 minutes.
    \end{tcolorbox}
    
    \caption{The instruction interface provided to human evaluators for the heritage dataset evaluation task. Above is the original Chinese version; below is the English translation.}
    \label{fig:human_eval_instructions}
\end{figure}

\begin{table}[htbp]
\centering
\scriptsize
\setlength{\tabcolsep}{6pt}
\renewcommand{\arraystretch}{1.15}
\begin{tabular}{ccccc}
\toprule
Question Type & Exact Agreement & Fleiss' $\kappa$ & Agreement Level \\
\midrule
Q1   & 56.0\% & 0.674 & High \\
Q2   & 64.0\% & 0.679 & High \\
Q3  & 58.0\% & 0.698 & High \\
Q4   & 16.0\% & 0.247 & Low \\
Q5   & 34.0\% & 0.486 & Moderate \\
Q6   & 52.0\% & 0.625 & High \\
Q7  & 38.0\% & 0.472 & Moderate \\
\midrule
Overall & 45.4\% & 0.592 & Moderate \\
\bottomrule
\end{tabular}
\caption{
Human answer agreement among three evaluators across question types. Exact agreement denotes the proportion of items for which all three evaluators gave the same answer, while Fleiss' $\kappa$ measures inter-evaluator answer consistency after accounting for chance agreement.
}
\label{tab:human_answer_agreement}
\end{table}

\section{Experiment Details}
\label{app:Experiment}
During the experiments, we used the \texttt{transformers} \cite{wolf-etal-2020-transformers} and \texttt{pytorch} \cite{NEURIPS2019_9015} library for deploying the models.
All experiments were conducted on NVIDIA A100 80GB GPUs. We used the official model implementations, processors, and recommended dependency settings for each VLM. All models were loaded in \texttt{bfloat16} precision and evaluated with deterministic decoding. Specifically, we used greedy decoding with \texttt{do\_sample=False}, without temperature, top-\(p\), or top-\(k\) sampling. Unless otherwise specified, \texttt{max\_new\_tokens} was set to 32.

\begin{table}[htbp]
\centering
\tiny
\begin{tabular}{lccc}
\toprule
Model & Decoding & Max tokens & Visual input setting \\
\midrule
CogVLM2 & Greedy & 32 & Default processor \\
DeepSeek-VL2 & Greedy & 32 & Official processor \\
GLM-4.6V-Flash & Constrained & 1 & Longest side limited to 512 px \\
InternVL2.5 & Greedy & 32 & \(448\times448\), up to 12 tiles \\
Qwen2.5-VL & Greedy & 32 & Official dynamic resizing \\
Qwen3-VL & Greedy & 32 & Official dynamic resizing \\
\bottomrule
\end{tabular}
\caption{Inference settings used for model evaluation.}
\label{tab:Inference_settings}
\end{table}

For model-specific settings showns in Table \ref{tab:Inference_settings}, Qwen3-VL was evaluated with \texttt{enable\_thinking=False}. GLM-4.6V-Flash used force-choice constrained decoding, where only valid option-letter tokens were allowed. Since CogVLM2 does not natively support the Q2 multi-image setting, the candidate images were concatenated into a single labeled grid image before inference.

\section{Additional Results}
\label{app:additional}

\begin{figure*}[htbp]
    \centering
    \includegraphics[width=\linewidth]{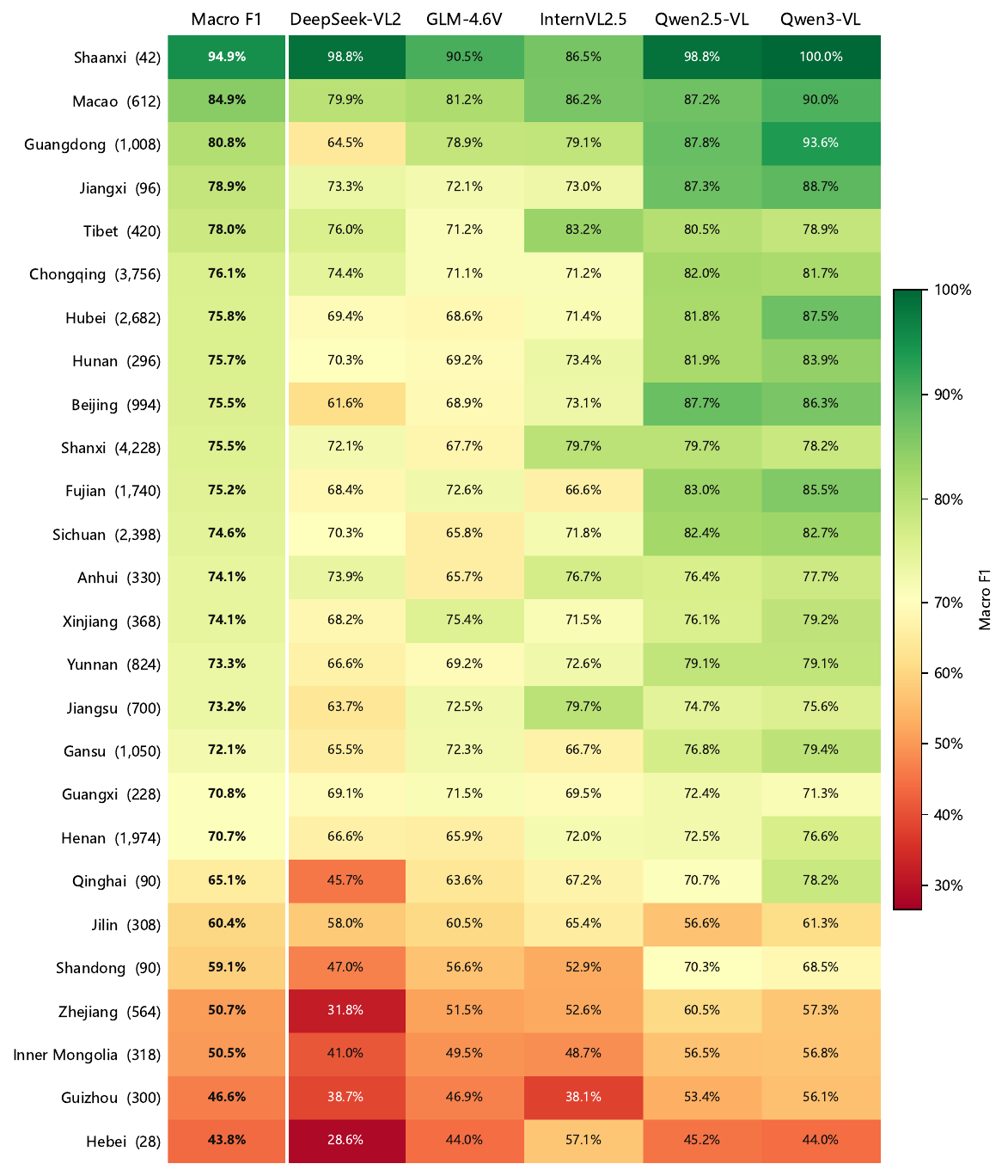}
    \caption{Province-level Macro-F1 for five VLMs (CogVLM2 excluded), sorted by the Macro F1 column (cross-model mean, separated by a white bar). Each cell is the macro-average of per-stratum F1 scores across all valid question-type × language combinations, giving equal weight to every question type independent of provincial item counts. Sample size per province is shown in parentheses.}
    \label{fig:province_f1_table}
\end{figure*}

Figure \ref{fig:province_f1_table} presents a detailed breakdown of model performance across all 26 provinces. Performance ranges from 100\% (Qwen3-VL on Shaanxi) to 28.6\% (DeepSeek-VL2 on Hebei), revealing substantial geographic variation. High-performing provinces (Shaanxi, Guangdong, Yunnan) cluster in the upper portion, while low-performing provinces (Hebei, Guizhou, Zhejiang) appear at the bottom. This province-level analysis confirms the regional grounding bias observed in Section \ref{sec:Analysis}: VLMs do not uniformly understand Chinese heritage across regions.

\section{Error Case}
\label{app:error}

\begin{figure}[ht] % 保持单栏
\centering
\begin{tcolorbox}[
        colback=gray!2!white, colframe=gray!70!black, 
        title={Error case of Q2}, 
        fonttitle=\small\bfseries, boxrule=0.5pt, sharp corners,
        left=6pt, right=6pt, top=4pt, bottom=4pt
    ]
\begin{CJK}{UTF8}{gbsn}
\footnotesize

\textbf{Question (CN):} 以下哪个图片可能是在龙门石窟拍摄的?

\vspace{0.8mm}
\textbf{Question (EN):} Which of the following images was likely taken at the Longmen Grottoes?

\vspace{1.2mm}
\textbf{Options:}

\vspace{0.6mm}
\centering
\setlength{\tabcolsep}{4pt}
\renewcommand{\arraystretch}{0.95}

\begin{tabular}{@{}ccc@{}}
\textbf{A} & \textbf{B} & \textbf{C} \\[-0.2mm]
\includegraphics[width=0.29\columnwidth,keepaspectratio]{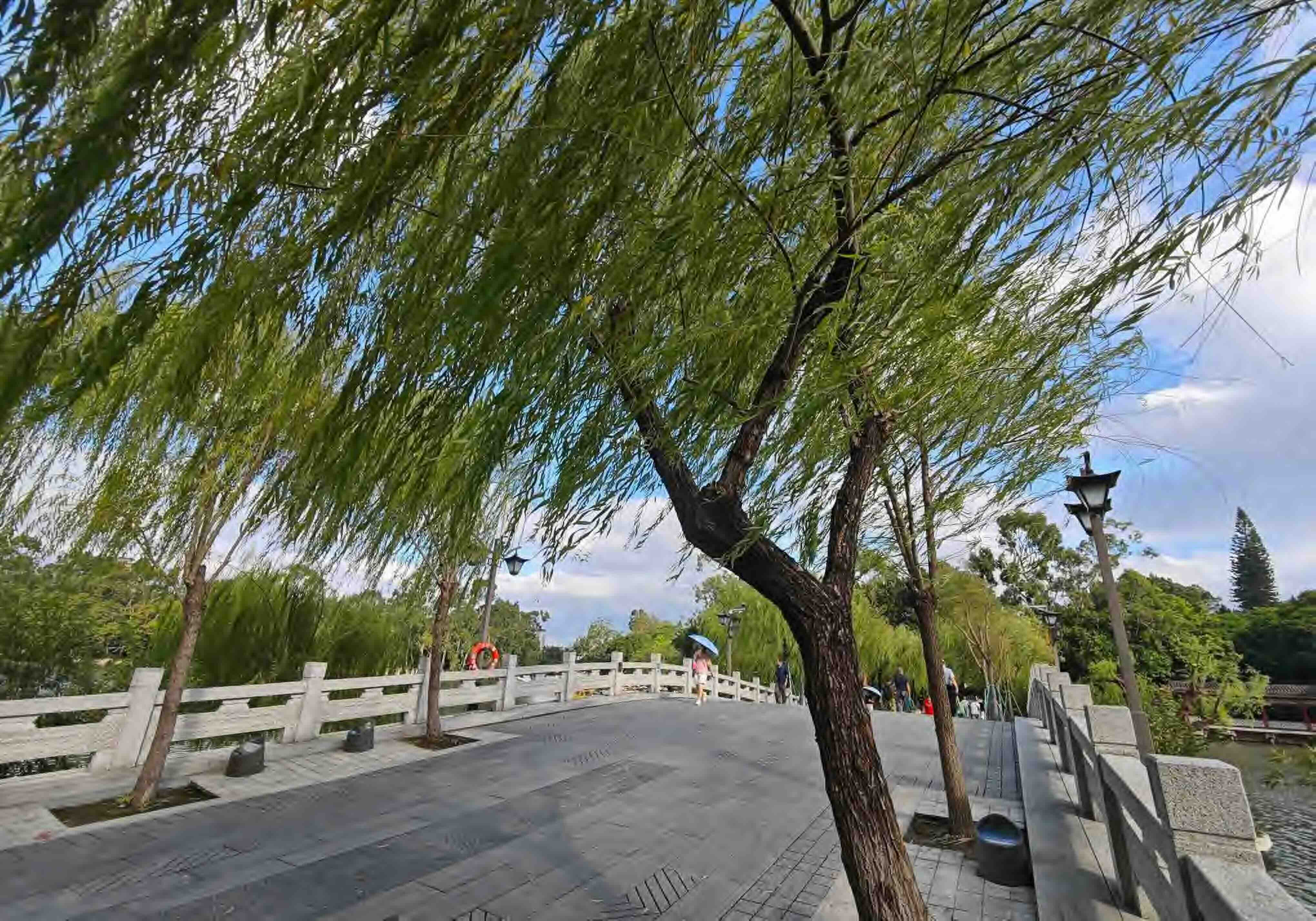} &
\includegraphics[width=0.29\columnwidth,keepaspectratio]{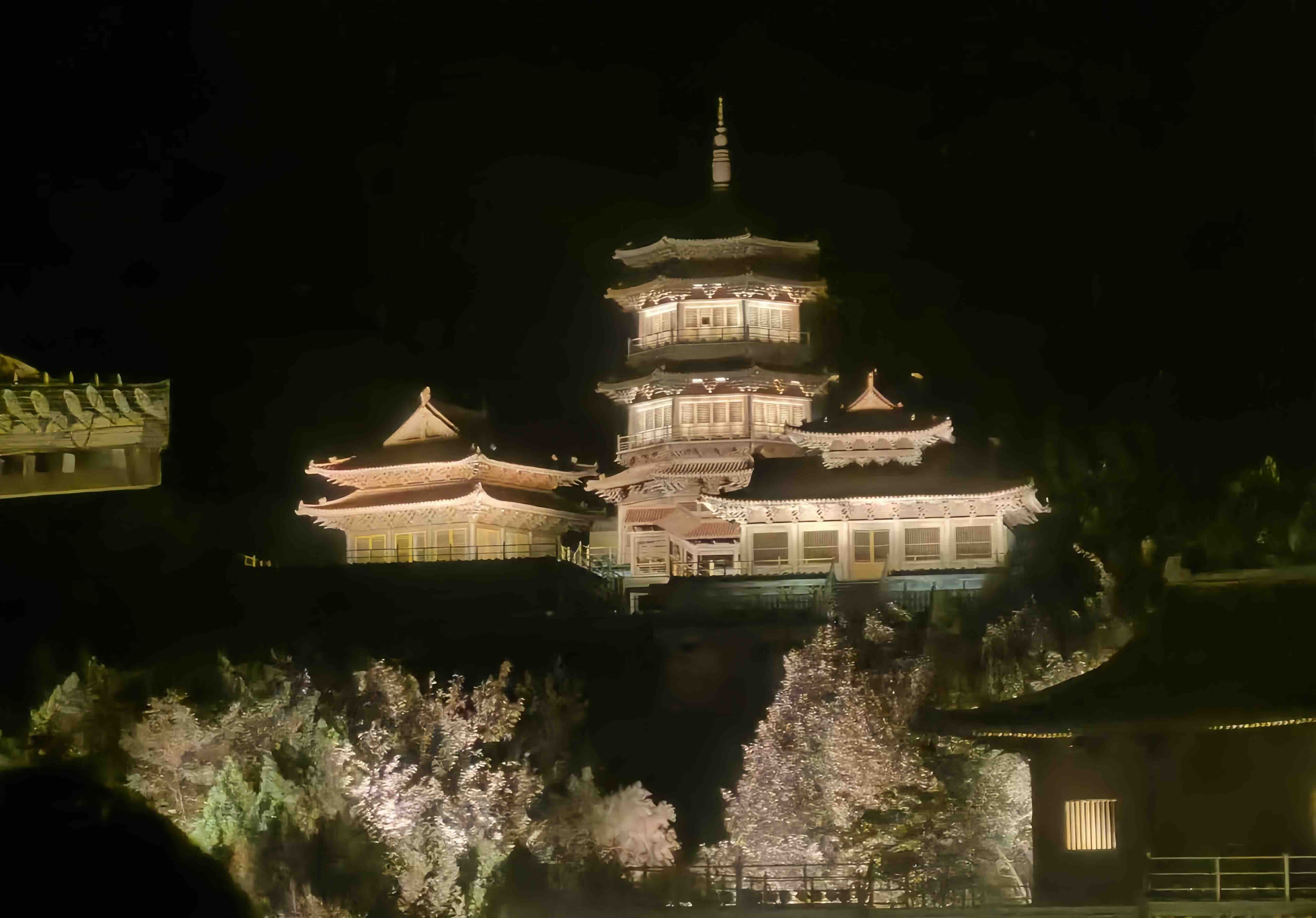} &
\includegraphics[width=0.29\columnwidth,keepaspectratio]{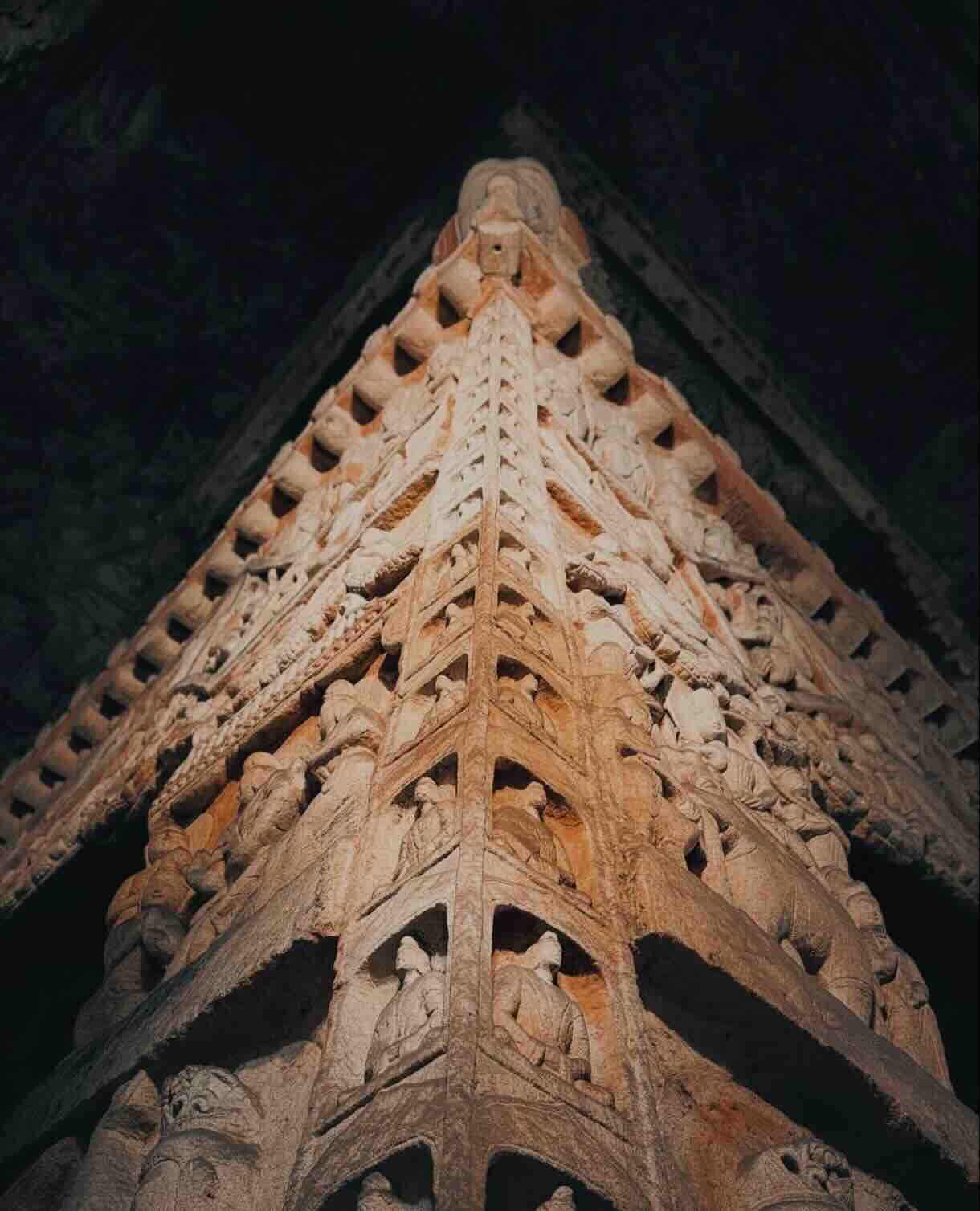} \\ \vspace{1mm}
\end{tabular}

\begin{tabular}{@{}cc@{}}
\textbf{D} & \textbf{E} \\[-0.2mm]
\includegraphics[width=0.29\columnwidth,keepaspectratio]{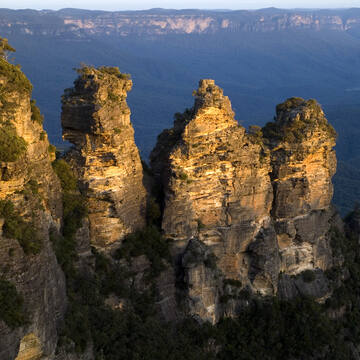} &
\includegraphics[width=0.29\columnwidth,keepaspectratio]{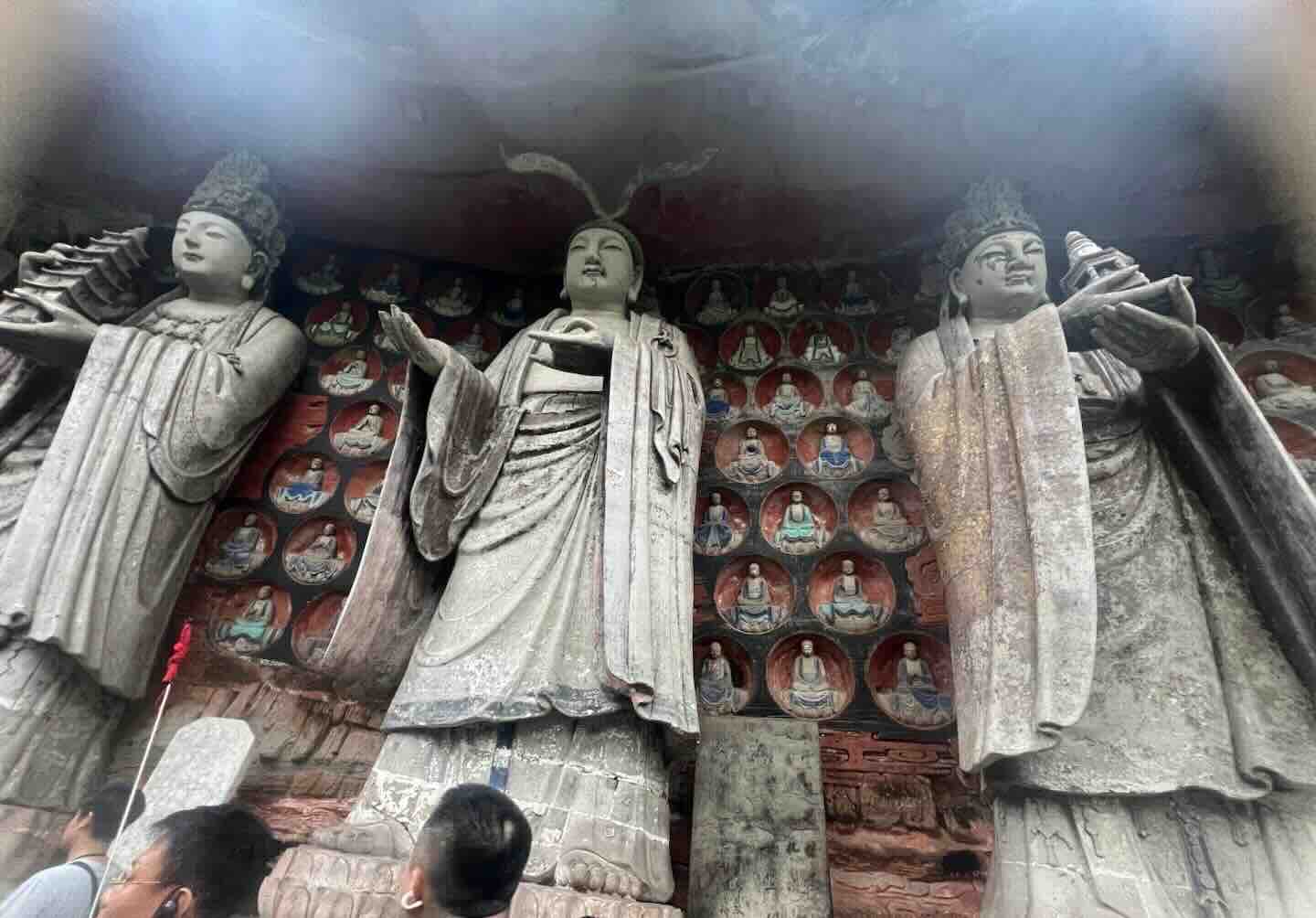}
\end{tabular}

\vspace{2mm}
\leftline{\footnotesize \textbf{Correct Answer:} C}
\leftline{\footnotesize \textbf{Models Anwser:} E}

\end{CJK}
\end{tcolorbox}
\begin{CJK}{UTF8}{gbsn}
\caption{
A representative failure case of Q2 Visual Grounding. The five candidates correspond to 
(A) 苏州古典园林 (Classical Gardens of Suzhou), 
(B) 登封 “天地之中”历史古迹 (Historic Monuments of Dengfeng in ``The Centre of Heaven and Earth''), 
(C) 龙门石窟 (Longmen Grottoes), 
(D) 蓝山国家公园 (Greater Blue Mountains Area, and 
(E) 大足石刻 (Dazu Rock Carvings). 
All models incorrectly select E despite C being correct, illustrating how similar heritage types lead to fine-grained discrimination failures.}
\label{fig:app_error_q2_example}
\end{CJK}
\end{figure}

Figure \ref{fig:app_error_q2_example} shows a typical same-type visual grounding failure. The model incorrectly selects Dazu Rock Carvings instead of Longmen Grottoes, although both sites share similar visual features such as Buddhist stone carvings, cliff-side niches, and sculptural figures. The error suggests that the model captures the coarse category of “Chinese Buddhist rock-carving heritage” but fails to ground the query to the correct site-specific visual identity. Rather than reflecting cross-cultural confusion, this case reveals an intra-cultural discrimination problem: the model relies on generic visual semantics such as “stone Buddha” or “grotto carving,” while missing finer spatial and architectural cues that distinguish Longmen from Dazu. This aligns with the overall Q2 error pattern, where same-type distractors account for the majority of wrong answers.

%------------------------------------
%           Q examples
%-----------------------------------
\section{Examples of Questions (Q1-Q7)}
\label{app:question_examples}

This appendix provides representative examples for the seven question types in ChinaHeritaQA, in Figures \ref{fig:app_q1_example}--\ref{fig:app_q7_example}.
All examples are evaluated under the shared system prompt in Appendix~\ref{app:prompts}. 
For readability, the shared prompt is omitted from each individual example. 
Each example includes the visual input, bilingual question text, five answer options, and the correct answer.

\begin{table*}[ht]
\centering
\small
\setlength{\tabcolsep}{4pt}
\renewcommand{\arraystretch}{1.15}
\begin{tabular}{p{0.18\textwidth} p{0.67\textwidth} c}
\toprule
\textbf{Negative Category} & \textbf{Prompt Set} & \textbf{Threshold} \\
\midrule

Food and drinks
&
\textit{a close-up photo of food};
\textit{a restaurant meal};
\textit{snacks and drinks};
\textit{street food};
\textit{a dining table with food};
\textit{a cup of coffee or tea};
\textit{a beverage bottle};
\textit{ice cream or dessert};
\textit{a menu with food};
\textit{a hotpot meal}
& 0.30 \\

\midrule

Tickets, maps, and brochures
&
\textit{an entrance ticket};
\textit{a tourist ticket};
\textit{a printed ticket};
\textit{a tourist map};
\textit{a travel map};
\textit{a guide brochure};
\textit{a folded brochure};
\textit{a paper travel guide};
\textit{a scenic area map board};
\textit{a museum guide leaflet}
& 0.30 \\

\midrule

Screenshots and user interfaces
&
\textit{a phone screenshot};
\textit{a social media screenshot};
\textit{a chat screenshot};
\textit{a website screenshot};
\textit{a map app screenshot};
\textit{a booking app screenshot};
\textit{a travel app screenshot};
\textit{a video thumbnail with user interface};
\textit{a screen capture};
\textit{a mobile app interface}
& 0.30 \\

\midrule

Text-heavy posters and advertisements
&
\textit{a poster with large text};
\textit{an advertisement poster};
\textit{a promotional banner};
\textit{a document with lots of text};
\textit{a photo with excessive text overlay};
\textit{a presentation slide};
\textit{a notice board full of text};
\textit{a signboard dominated by text};
\textit{a printed announcement};
\textit{a billboard advertisement}
& 0.30 \\

\midrule

Souvenirs and products
&
\textit{souvenirs on a shelf};
\textit{a souvenir shop product};
\textit{tourist merchandise};
\textit{a postcard};
\textit{a fridge magnet};
\textit{a keychain souvenir};
\textit{a toy model};
\textit{a cultural creative product};
\textit{a packaged gift};
\textit{a product display shelf}
& 0.30 \\

\midrule

Shops, markets, and restaurants
&
\textit{inside a souvenir shop};
\textit{inside a gift shop};
\textit{inside a restaurant};
\textit{inside a cafe};
\textit{a shopping street storefront};
\textit{a market stall};
\textit{a store interior};
\textit{a commercial shop interior};
\textit{a restaurant counter};
\textit{a food stall}
& 0.30 \\

\midrule

Hotels and transportation
&
\textit{a hotel room};
\textit{a bedroom in a hotel};
\textit{a train station};
\textit{an airport terminal};
\textit{inside an airplane};
\textit{inside a train};
\textit{inside a bus};
\textit{a car interior};
\textit{a parking lot};
\textit{a suitcase at a hotel}
& 0.30 \\

\midrule

Documents and QR codes
&
\textit{a QR code};
\textit{a payment code};
\textit{an ID card};
\textit{a receipt};
\textit{an invoice};
\textit{a travel itinerary document};
\textit{a printed form};
\textit{a certificate};
\textit{a boarding pass};
\textit{a reservation confirmation}
& 0.30 \\

\bottomrule
\end{tabular}
\caption{Negative prompt categories and forced-removal thresholds used in the CLIP-based semantic filtering step. The filter is designed to remove only high-confidence irrelevant social-media images. Categories that may still contain valid heritage evidence, such as tourists, crowds, selfies, night scenes, and architectural details, are not included in the forced-removal prompt set.}
\label{tab:clip_negative_prompts}
\end{table*}

% ----- Q1 ----------
\begin{figure*}[t]
\centering
\begin{tcolorbox}[
    enhanced,
    width=0.96\textwidth,
    colback=white,
    colframe=black!35,
    colbacktitle=black!55,
    coltitle=white,
    title={Q1 Identity Recognition},
    fonttitle=\bfseries,
    boxrule=0.5pt,
    arc=2mm,
    left=2mm,
    right=2mm,
    top=2mm,
    bottom=2mm
]
\begin{minipage}[c]{0.29\textwidth}
    \centering
    \includegraphics[
        width=\linewidth,
        height=0.23\textheight,
        keepaspectratio
    ]{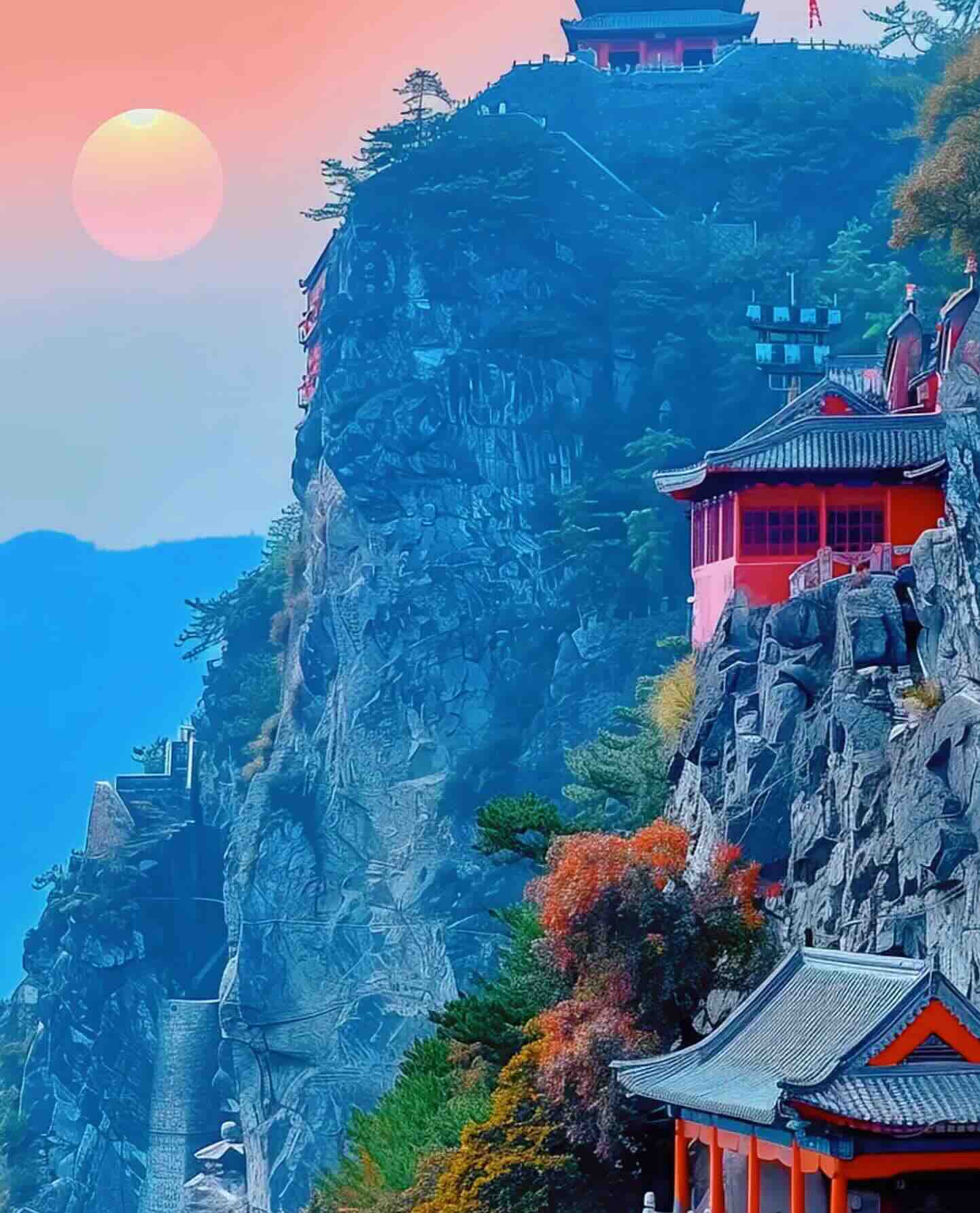}

    \vspace{1mm}
    {\footnotesize \textbf{Correct Answer:} A}
\end{minipage}
\hfill
\begin{minipage}[c]{0.63\textwidth}
\footnotesize
\renewcommand{\arraystretch}{1.05}
\setlength{\tabcolsep}{3pt}

\begin{CJK}{UTF8}{gbsn}
\textbf{Question (CN):} 图片中展示的是以下哪处文化或自然遗产地？

\vspace{0.8mm}
\textbf{Question (EN):} Which of the following cultural or natural heritage sites is depicted in this image?

\vspace{1mm}
\textbf{Options:}
\vspace{0.5mm}

\begin{tabularx}{\linewidth}{@{}>{\bfseries}l >{\raggedright\arraybackslash}X >{\raggedright\arraybackslash}X@{}}
 & \textbf{Chinese} & \textbf{English} \\
A. & 武当山古建筑群 
   & Ancient Building Complex in the Wudang Mountains \\
B. & 武陵源风景名胜区 
   & Wulingyuan Scenic and Historic Interest Area \\
C. & 湖北神农架 
   & Hubei Shennongjia \\
D. & 泰山 
   & Mount Taishan \\
E. & 圣米歇尔山及其海湾 
   & Mont-Saint-Michel and its Bay \\
\end{tabularx}
\end{CJK}
\end{minipage}
\end{tcolorbox}

\caption{
Representative example of Q1 Identity Recognition in ChinaHeritaQA. 
The left side shows the visual input, while the right side presents the bilingual question stems, bilingual answer options, and the gold answer.
}
\label{fig:app_q1_example}
\end{figure*}

% ----- Q2 ----------
\begin{figure*}[t]
\centering
\begin{tcolorbox}[
    enhanced,
    width=0.96\textwidth,
    colback=white,
    colframe=black!35,
    colbacktitle=black!55,
    coltitle=white,
    title={Q2 Visual Grounding},
    fonttitle=\bfseries,
    boxrule=0.5pt,
    arc=2mm,
    left=2mm,
    right=2mm,
    top=2mm,
    bottom=2mm
]
\begin{CJK}{UTF8}{gbsn}
\footnotesize

\textbf{Question (CN):} 以下哪张图片可能是在武当山古建筑群拍摄的？

\vspace{0.8mm}
\textbf{Question (EN):} Which of the following images was likely taken at the Ancient Building Complex in the Wudang Mountains?

\vspace{1.2mm}
\textbf{Options:}

\vspace{0.6mm}
\centering
\setlength{\tabcolsep}{2pt}
\renewcommand{\arraystretch}{0.95}

\begin{tabular}{@{}ccccc@{}}
\textbf{A} & \textbf{B} & \textbf{C} & \textbf{D} & \textbf{E} \\[-0.2mm]
\includegraphics[width=0.175\textwidth,height=0.115\textheight,keepaspectratio]{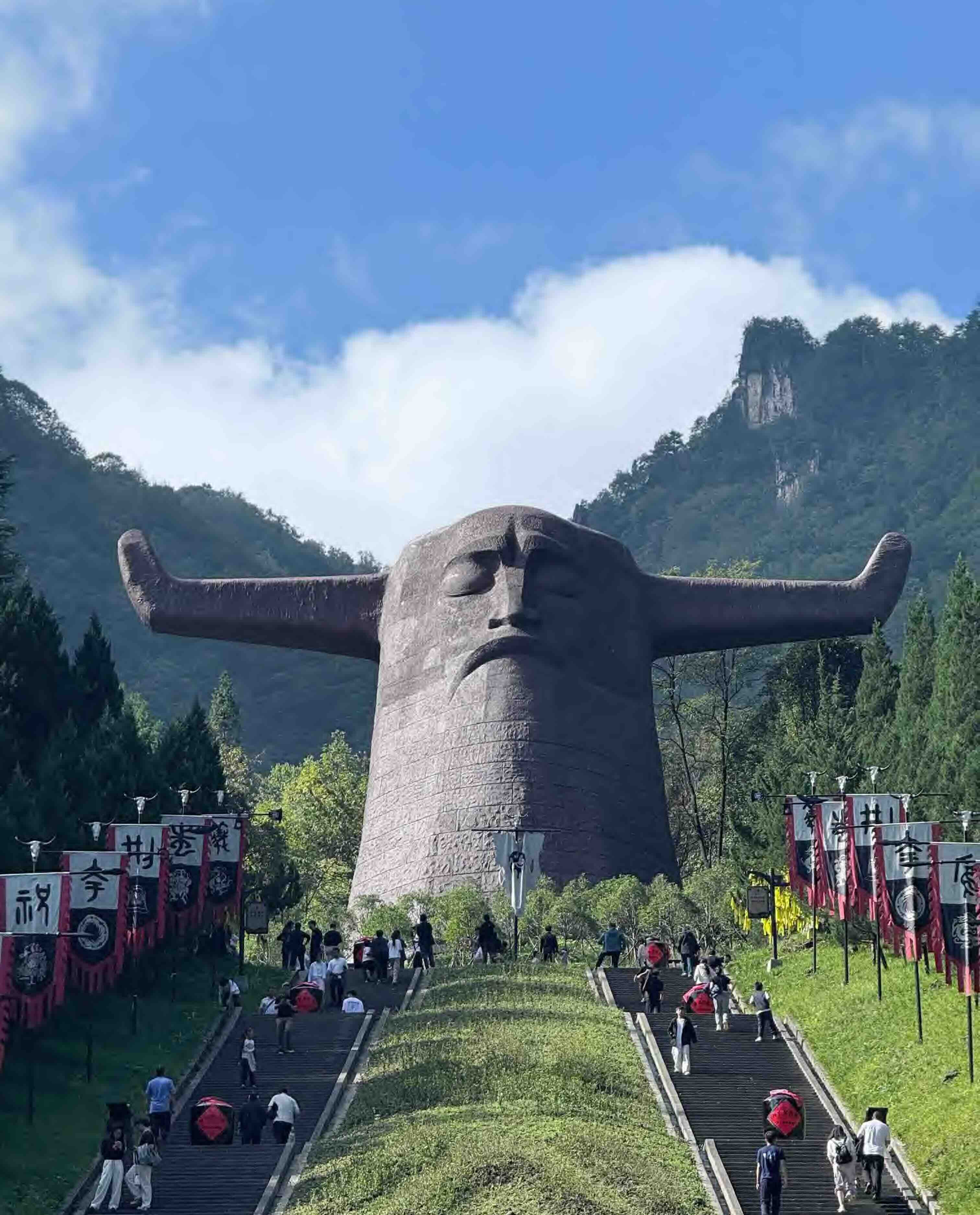} &
\includegraphics[width=0.175\textwidth,height=0.115\textheight,keepaspectratio]{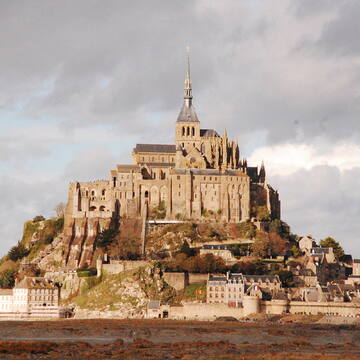} &
\includegraphics[width=0.175\textwidth,height=0.115\textheight,keepaspectratio]{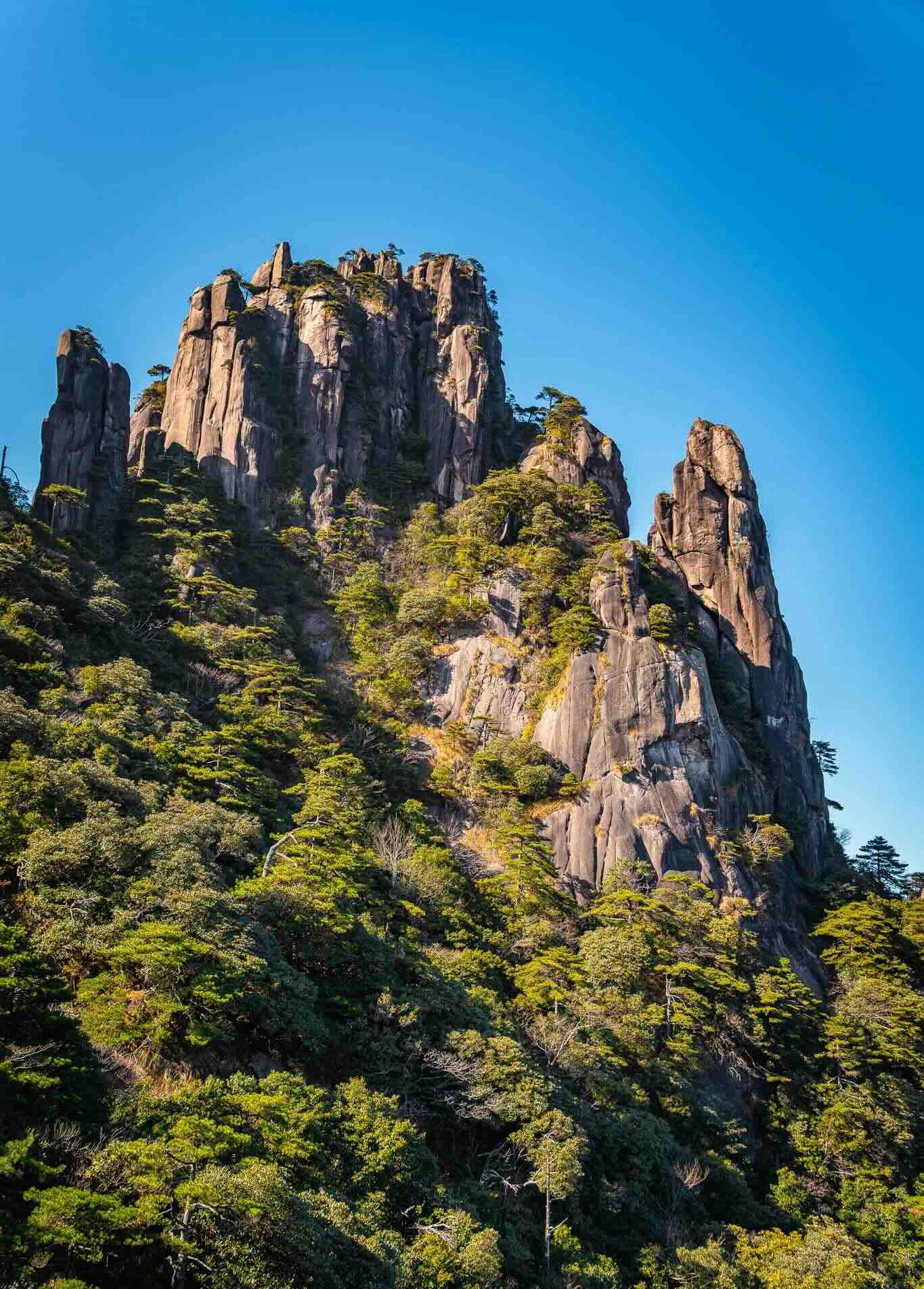} &
\includegraphics[width=0.175\textwidth,height=0.115\textheight,keepaspectratio]{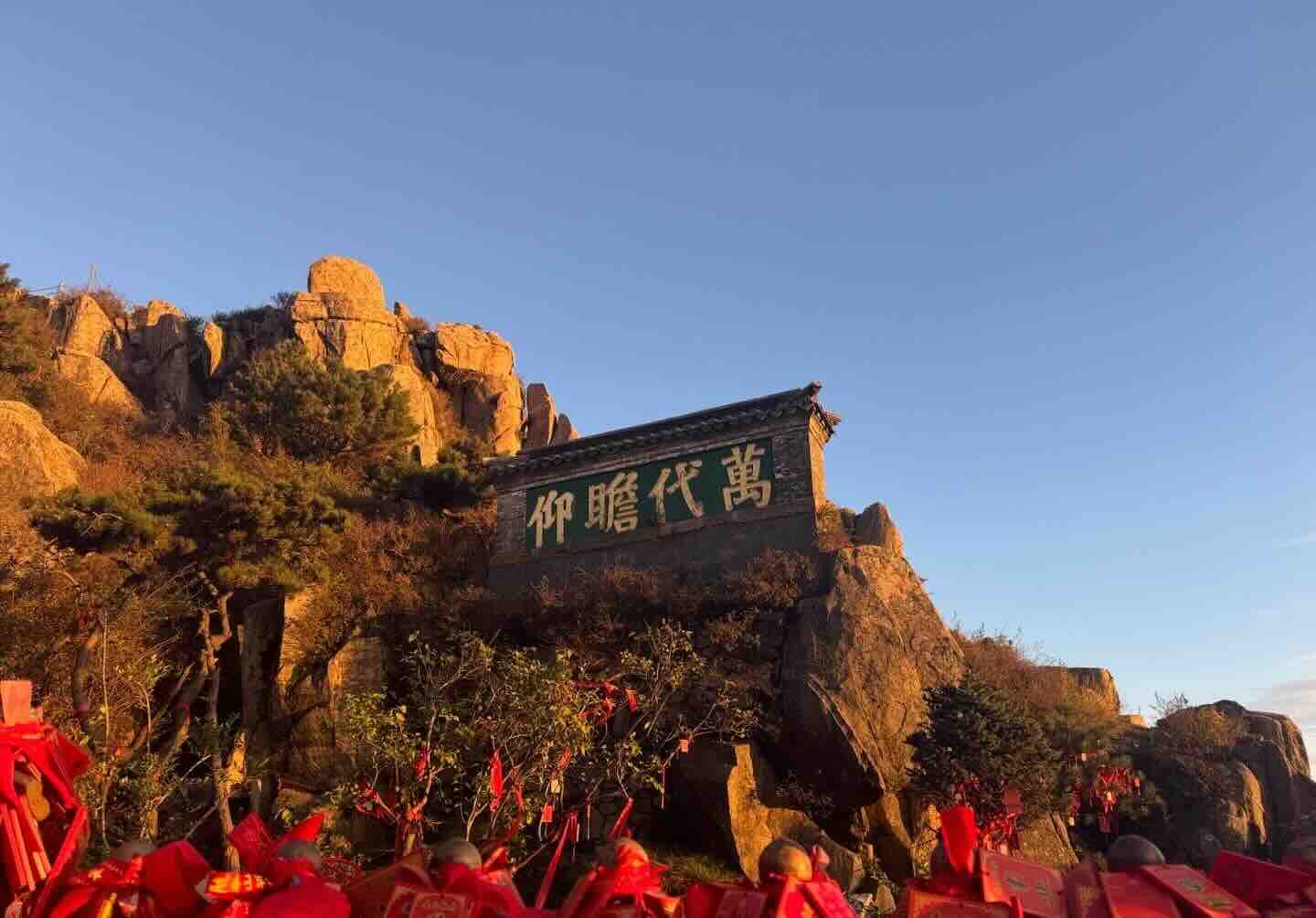} &
\includegraphics[width=0.175\textwidth,height=0.115\textheight,keepaspectratio]{Figures/appendix/Q1.jpeg}
\end{tabular}

\vspace{1mm}
{\footnotesize \textbf{Correct Answer:} E}

\end{CJK}
\end{tcolorbox}

\caption{
Representative example of Q2 Visual Grounding in ChinaHeritaQA. 
The model is given a heritage-site name and must select the corresponding image from multiple visual candidates.
}
\label{fig:app_q2_example}
\end{figure*}

% ----- Q3 ----------
\begin{figure*}[t]
\centering
\begin{tcolorbox}[
    enhanced,
    width=0.96\textwidth,
    colback=white,
    colframe=black!35,
    colbacktitle=black!55,
    coltitle=white,
    title={Q3 Description Matching},
    fonttitle=\bfseries,
    boxrule=0.5pt,
    arc=2mm,
    left=1.8mm,
    right=1.8mm,
    top=1.8mm,
    bottom=1.8mm
]

% left image block
\begin{minipage}[t]{0.25\textwidth}
    \vspace{0pt}
    \centering
    \includegraphics[
        width=1\linewidth,
        height=0.19\textheight,
        keepaspectratio
    ]{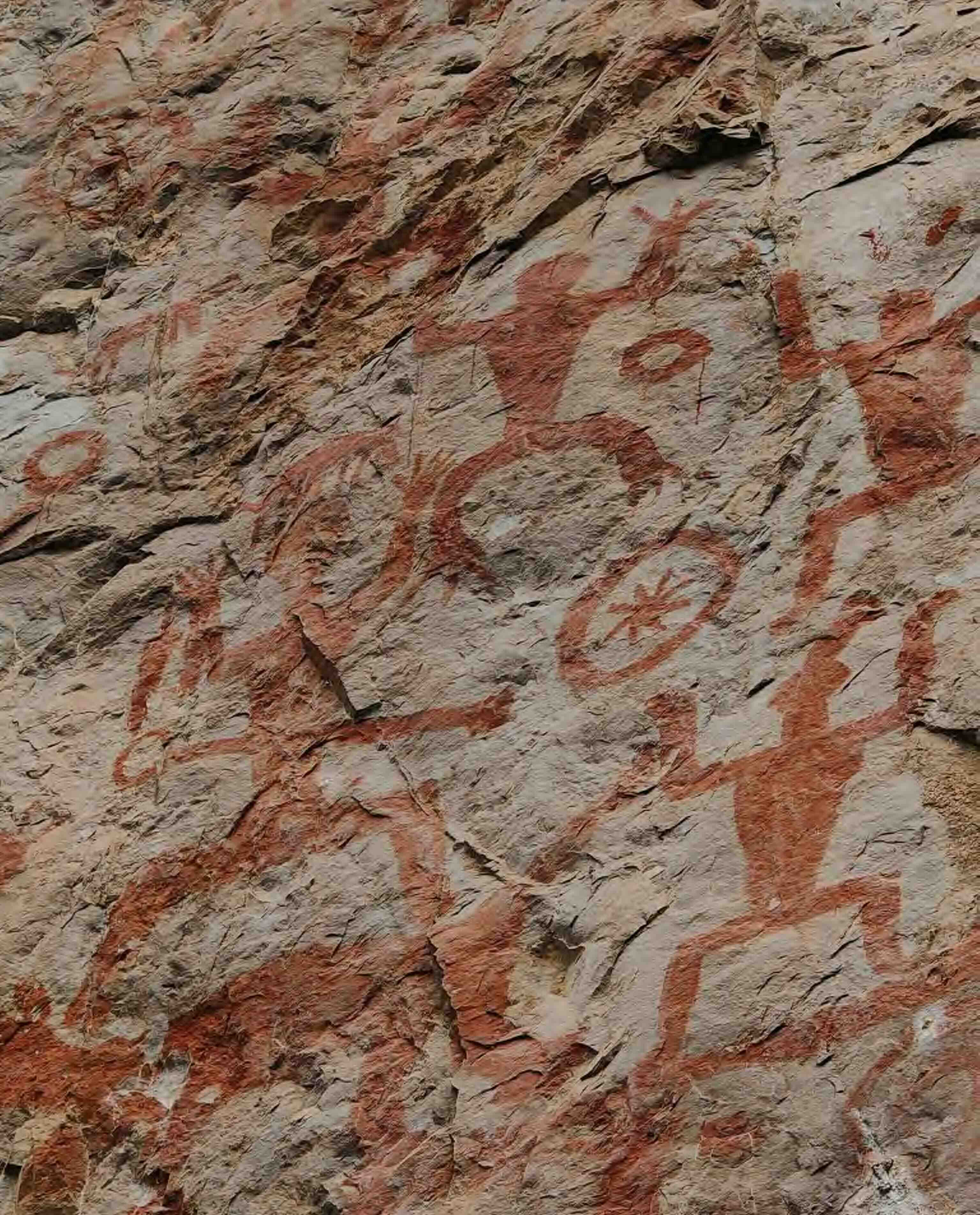}

    \vspace{1mm}
    {\footnotesize \textbf{Correct Answer:} E}
\end{minipage}
\hfill
% right question block
\begin{minipage}[t]{0.70\textwidth}
\vspace{0pt}
\scriptsize
\renewcommand{\arraystretch}{0.9}
\setlength{\tabcolsep}{2pt}

\begin{CJK}{UTF8}{gbsn}

\textbf{Question (CN):} 关于该图片简要介绍正确的是？

\vspace{0.5mm}
\textbf{Question (EN):} Which brief introduction regarding this picture is correct?

\vspace{0.7mm}
\textbf{Options:}
\vspace{0.3mm}

\begin{tabularx}{\linewidth}{
@{}
>{\bfseries}p{0.025\linewidth}
>{\raggedright\arraybackslash}p{0.39\linewidth}
>{\raggedright\arraybackslash}p{0.53\linewidth}
@{}
}
 & \textbf{Chinese} & \textbf{English} \\

A. & 以近代国际社区的街区、住宅、公共设施和多样建筑风格为特征，反映通商背景下多元文化共处和现代城市生活。
   & Marked by neighbourhoods, residences, public facilities, and varied architectural styles of a modern international settlement, the property reflects multicultural coexistence and urban life shaped by treaty-port history. \\

B. & 以山地梯田、村寨、森林和水系共同构成的农业文化景观，体现少数民族社群长期适应山地环境的水土管理与生活秩序。
   & This agricultural cultural landscape combines mountain terraces, villages, forests, and water systems, showing how an ethnic community has long adapted to upland conditions through water-soil management and social organization. \\

C. & 兼具山地自然景观、历史建筑、宗教遗存和近代别墅群，体现山水审美、文化交流和避暑休闲功能的叠加。
   & Combining mountain scenery, historic buildings, religious remains, and modern villas, the site reflects layered functions of landscape appreciation, cultural exchange, and summer retreat. \\

D. & 以多层防御性乡村住宅和周边村落为特征，融合中西建筑与装饰元素，反映侨乡社会、家族防卫和近代跨文化联系。
   & Characterized by multi-storey defensive rural residences and surrounding villages, the property blends Chinese and Western architectural and decorative forms, reflecting emigrant communities, family defense, and modern cross-cultural links. \\

E. & 以江岸崖壁上的岩画与周边山水共同构成文化景观，反映古代族群的仪式活动、图像表达和青铜鼓文化传统。
   & The cultural landscape combines cliff-side rock paintings with the surrounding river-and-mountain setting, reflecting ritual activity, image-making, and bronze-drum cultural traditions of an ancient community. \\

\end{tabularx}
\end{CJK}

\end{minipage}
\end{tcolorbox}

\caption{
Representative example of Q3 Description Matching in ChinaHeritaQA. 
The model must choose the correct bilingual descriptive summary according to the visual evidence.
}
\label{fig:app_q3_example}
\end{figure*}

% ----- Q4 ----------
\begin{figure*}[t]
\centering
\begin{tcolorbox}[
    enhanced,
    width=0.96\textwidth,
    colback=white,
    colframe=black!35,
    colbacktitle=black!55,
    coltitle=white,
    title={Q4 Historical Periodization},
    fonttitle=\bfseries,
    boxrule=0.5pt,
    arc=2mm,
    left=2mm,
    right=2mm,
    top=2mm,
    bottom=2mm
]
\begin{minipage}[c]{0.29\textwidth}
    \centering
    \includegraphics[
        width=\linewidth,
        height=0.23\textheight,
        keepaspectratio
    ]{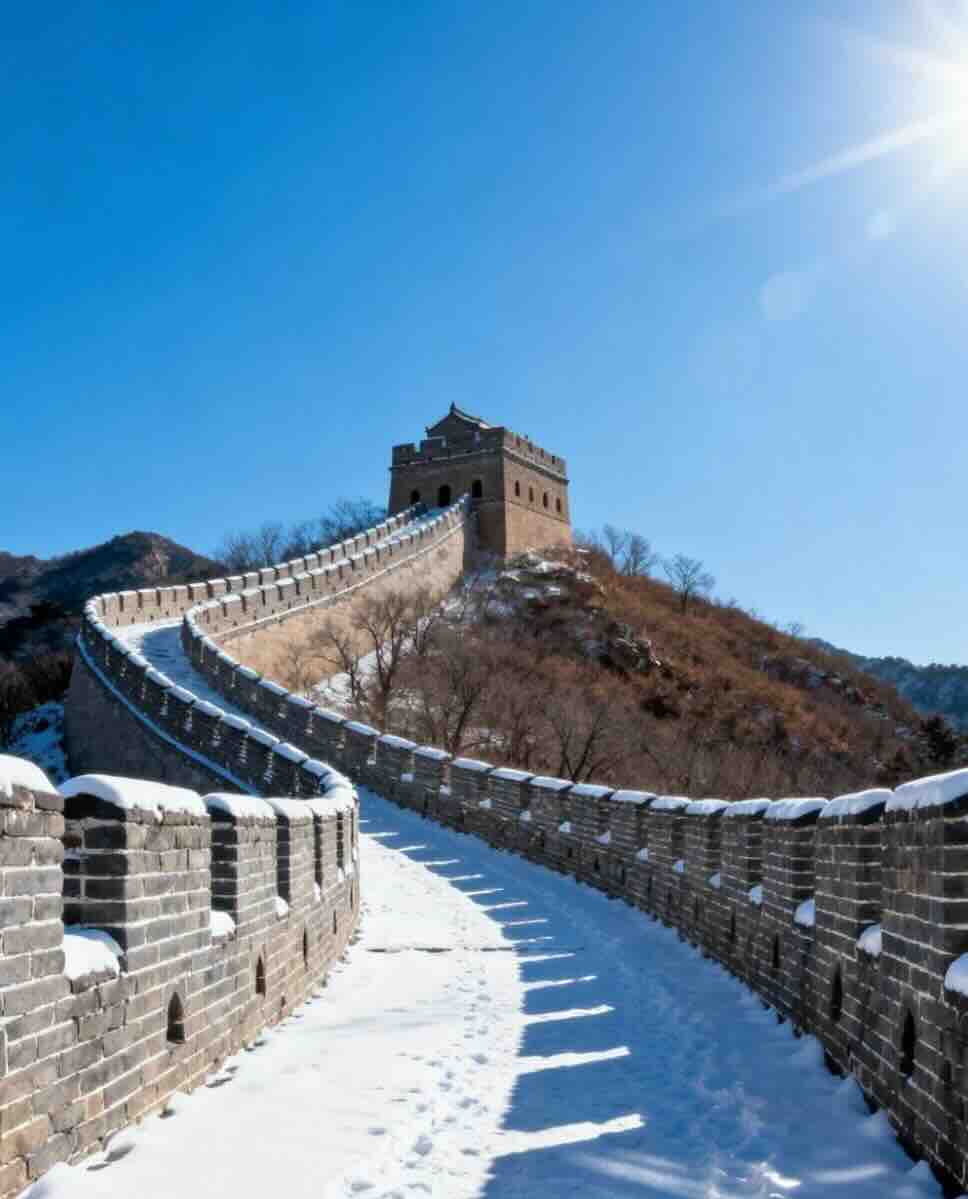}

    \vspace{1mm}
    {\footnotesize \textbf{Correct Answer:} C}
\end{minipage}
\hfill
\begin{minipage}[c]{0.63\textwidth}
\footnotesize
\renewcommand{\arraystretch}{1.05}
\setlength{\tabcolsep}{3pt}

\begin{CJK}{UTF8}{gbsn}
\textbf{Question (CN):} 该图片中的建筑群可能建于哪个朝代?

\vspace{0.8mm}
\textbf{Question (EN):} In which dynasty might the building complex in this picture have been built?

\vspace{1mm}
\textbf{Options:}
\vspace{0.5mm}

\begin{tabularx}{\linewidth}{@{}>{\bfseries}l >{\raggedright\arraybackslash}X >{\raggedright\arraybackslash}X@{}}
 & \textbf{Chinese} & \textbf{English} \\
A. & 清朝 & Qing Dynasty \\
B. & 欧盟一体化时代 & The era of EU integration \\
C. & 秦朝 & Qin Dynasty \\
D. & 南北朝时期 & The Northern and Southern Dynasties period \\
E. & 五代十国时期 & The Five Dynasties and Ten Kingdoms period \\
\end{tabularx}
\end{CJK}
\end{minipage}
\end{tcolorbox}

\caption{
Representative example of Q4 Historical Periodization in ChinaHeritaQA. 
The model must infer the possible historical period or dynasty associated with the visual input.
}
\label{fig:app_q4_example}
\end{figure*}

% ----- Q5 ----------
\begin{figure*}[t]
\centering
\begin{tcolorbox}[
    enhanced,
    width=0.96\textwidth,
    colback=white,
    colframe=black!35,
    colbacktitle=black!55,
    coltitle=white,
    title={Q5 Historical Contextualization},
    fonttitle=\bfseries,
    boxrule=0.5pt,
    arc=2mm,
    left=2mm,
    right=2mm,
    top=2mm,
    bottom=2mm
]
\begin{minipage}[c]{0.29\textwidth}
    \centering
    \includegraphics[
        width=\linewidth,
        height=0.23\textheight,
        keepaspectratio
    ]{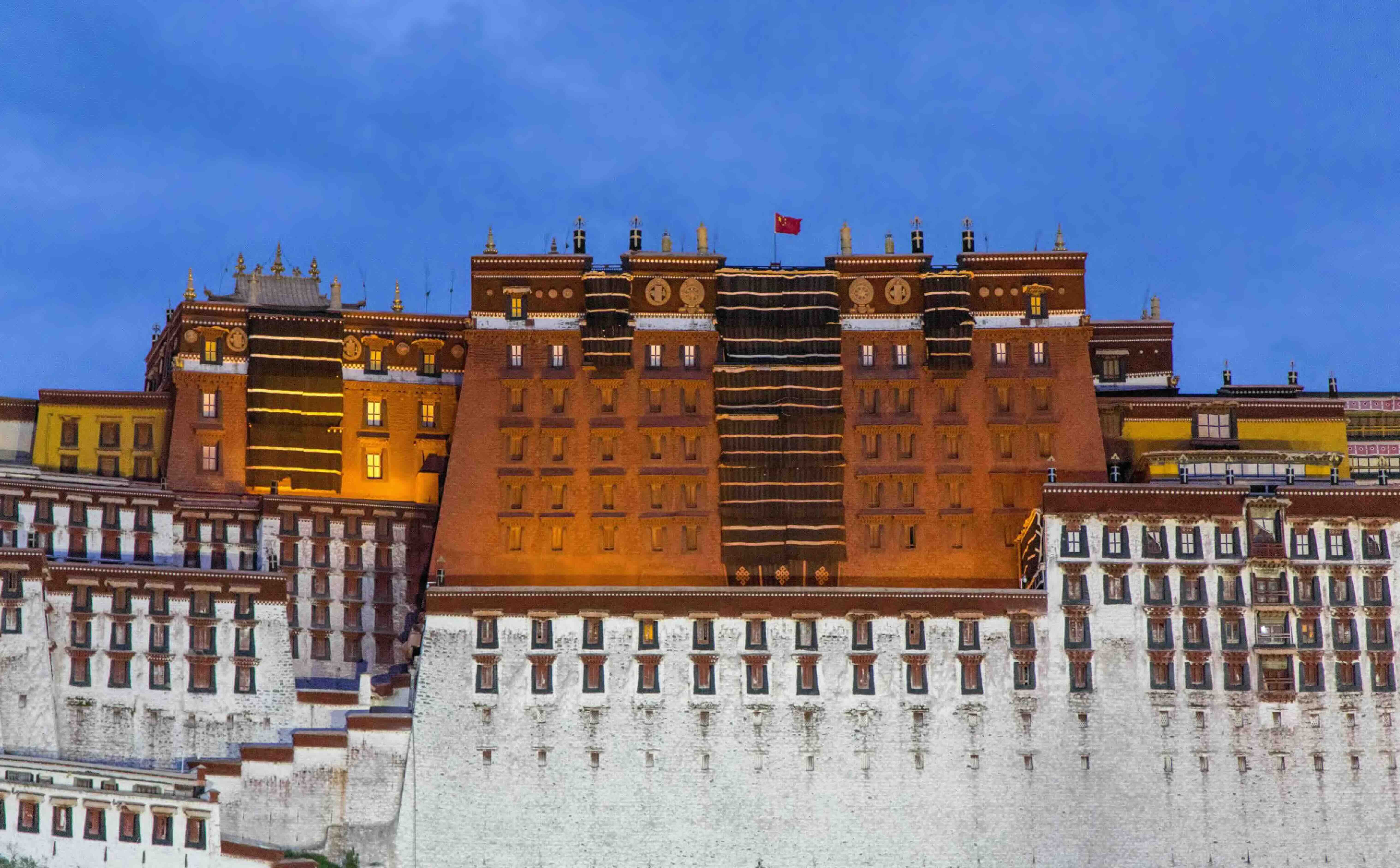}

    \vspace{1mm}
    {\footnotesize \textbf{Correct Answer:} E}
\end{minipage}
\hfill
\begin{minipage}[c]{0.63\textwidth}
\footnotesize
\renewcommand{\arraystretch}{1.05}
\setlength{\tabcolsep}{3pt}

\begin{CJK}{UTF8}{gbsn}
\textbf{Question (CN):} 关于该图片历史背景介绍正确的是?

\vspace{0.8mm}
\textbf{Question (EN):} Which description of the historical background of this picture is correct?

\vspace{1mm}
\textbf{Options:}
\vspace{0.5mm}

\begin{tabularx}{\linewidth}{@{}>{\bfseries}l >{\raggedright\arraybackslash}X >{\raggedright\arraybackslash}X@{}}
 & \textbf{Chinese} & \textbf{English} \\
A. & 北魏至唐代开凿，艺术风格演变
   & Carved from the Northern Wei to the Tang Dynasty, the artistic style evolved \\
B. & 自公元前3世纪起建设的防御工程
   & Defensive works built since the 3rd century BC \\
C. & 已有超过4万年的连续居住历史
   & It has a continuous inhabited history of over 40,000 years. \\
D. & 公元9至13世纪岩刻艺术遗存
   & Rock carvings dating from the 9th to 13th centuries \\
E. & 始建于公元7世纪，象征藏传佛教地位
   & Built in the 7th century AD, it symbolizes the status of Tibetan Buddhism. \\
\end{tabularx}
\end{CJK}
\end{minipage}
\end{tcolorbox}

\caption{
Representative example of Q5 Historical Contextualization in ChinaHeritaQA. 
The model must select the correct historical background associated with the depicted heritage site.
}
\label{fig:app_q5_example}
\end{figure*}

% ----- Q6 ----------
\begin{figure*}[t]
\centering
\begin{tcolorbox}[
    enhanced,
    width=0.96\textwidth,
    colback=white,
    colframe=black!35,
    colbacktitle=black!55,
    coltitle=white,
    title={Q6 Functional Analysis},
    fonttitle=\bfseries,
    boxrule=0.5pt,
    arc=2mm,
    left=2mm,
    right=2mm,
    top=2mm,
    bottom=2mm
]
\begin{minipage}[c]{0.29\textwidth}
    \centering
    \includegraphics[
        width=\linewidth,
        height=0.23\textheight,
        keepaspectratio
    ]{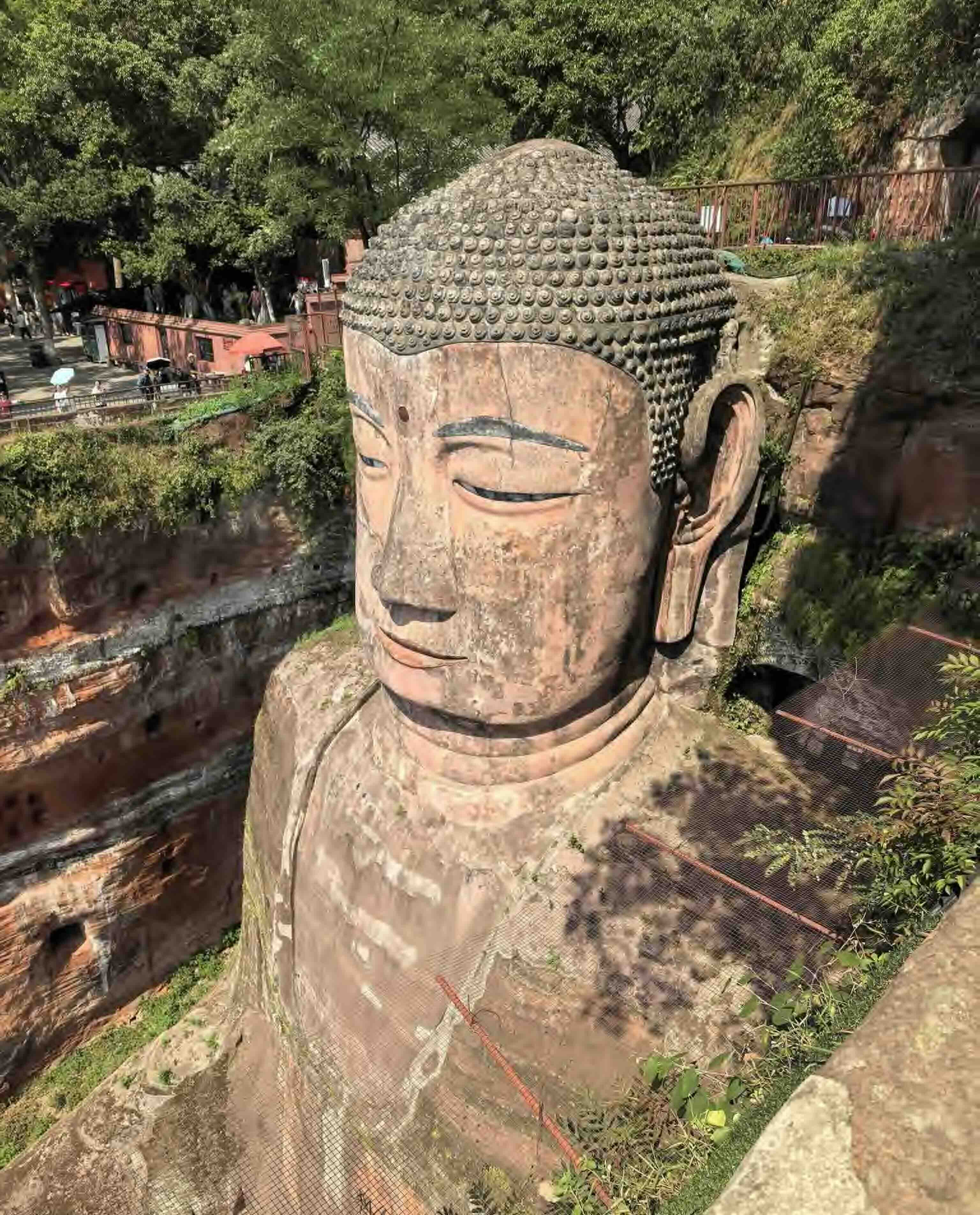}

    \vspace{1mm}
    {\footnotesize \textbf{Correct Answer:} C}
\end{minipage}
\hfill
\begin{minipage}[c]{0.63\textwidth}
\footnotesize
\renewcommand{\arraystretch}{1.05}
\setlength{\tabcolsep}{3pt}

\begin{CJK}{UTF8}{gbsn}
\textbf{Question (CN):} 关于该图片主要的功能介绍正确的是?

\vspace{0.8mm}
\textbf{Question (EN):} Which description of the main function of this picture is correct?

\vspace{1mm}
\textbf{Options:}
\vspace{0.5mm}

\begin{tabularx}{\linewidth}{@{}>{\bfseries}l >{\raggedright\arraybackslash}X >{\raggedright\arraybackslash}X@{}}
 & \textbf{Chinese} & \textbf{English} \\
A. & 防御与聚落居住
   & Defense and settlement \\
B. & 生态保护与科学研究
   & Ecological protection and scientific research \\
C. & 佛教圣地与文化景观
   & Buddhist holy sites and cultural landscapes \\
D. & 自然景观旅游
   & Natural Landscape Tourism \\
E. & 自然风光观赏与植物保护
   & Natural scenery appreciation and plant protection \\
\end{tabularx}
\end{CJK}
\end{minipage}
\end{tcolorbox}

\caption{
Representative example of Q6 Functional Analysis in ChinaHeritaQA. 
The model must infer the main function of the depicted heritage site from visual and cultural cues.
}
\label{fig:app_q6_example}
\end{figure*}

% ----- Q7 ----------
\begin{figure*}[t]
\centering
\begin{tcolorbox}[
    enhanced,
    width=0.96\textwidth,
    colback=white,
    colframe=black!35,
    colbacktitle=black!55,
    coltitle=white,
    title={Q7 Architectural Analysis},
    fonttitle=\bfseries,
    boxrule=0.5pt,
    arc=2mm,
    left=2mm,
    right=2mm,
    top=2mm,
    bottom=2mm
]
\begin{minipage}[c]{0.29\textwidth}
    \centering
    \includegraphics[
        width=\linewidth,
        height=0.23\textheight,
        keepaspectratio
    ]{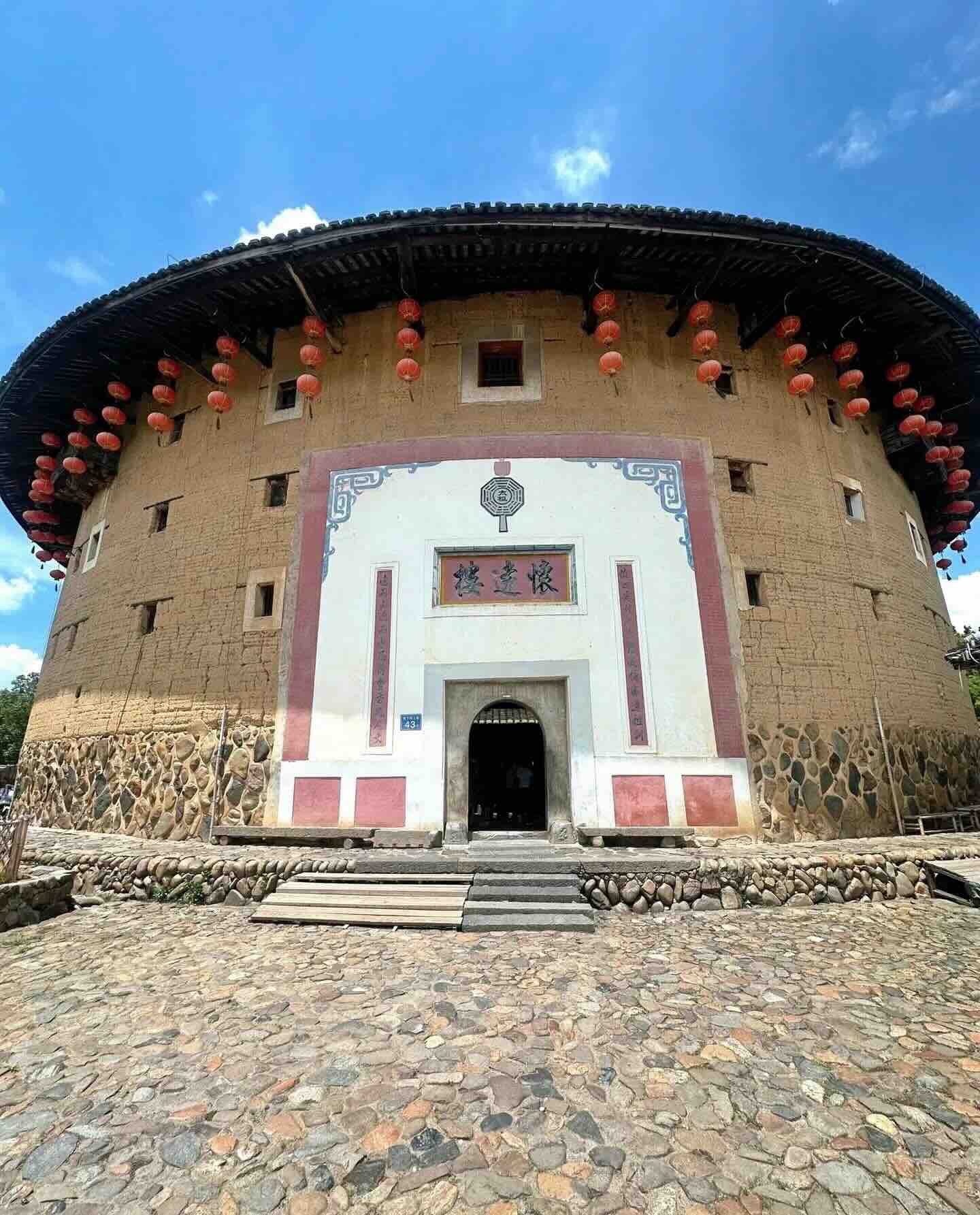}

    \vspace{1mm}
    {\footnotesize \textbf{Correct Answer:} A}
\end{minipage}
\hfill
\begin{minipage}[c]{0.63\textwidth}
\footnotesize
\renewcommand{\arraystretch}{1.05}
\setlength{\tabcolsep}{3pt}

\begin{CJK}{UTF8}{gbsn}
\textbf{Question (CN):} 关于该图片建筑用途介绍正确的是?

\vspace{0.8mm}
\textbf{Question (EN):} Which description of the architectural usage of this picture is correct?

\vspace{1mm}
\textbf{Options:}
\vspace{0.5mm}

\begin{tabularx}{\linewidth}{@{}>{\bfseries}l >{\raggedright\arraybackslash}X >{\raggedright\arraybackslash}X@{}}
 & \textbf{Chinese} & \textbf{English} \\
A. & 居民聚居与防御
   & Residential settlement and defense \\
B. & 抵御外敌入侵，传递军事信息
   & Resisting foreign invasion and transmitting military information \\
C. & 本笃会修道院
   & Benedictine Monastery \\
D. & 寺庙、书院和遗址供学术与宗教活动
   & Temples, academies, and ruins served for academic and religious activities. \\
E. & 古城墙、民居、寺庙等综合用途
   & Ancient city walls, residences, temples and other multi-purpose buildings \\
\end{tabularx}
\end{CJK}
\end{minipage}
\end{tcolorbox}

\caption{
Representative example of Q7 Architectural Analysis in ChinaHeritaQA. 
The model must identify the correct architectural usage of the depicted heritage site.
}
\label{fig:app_q7_example}
\end{figure*}

\end{document}